\documentclass[a4paper,fleqn]{cas-dc}

\usepackage[authoryear]{natbib}
\usepackage{url}
\usepackage[most]{tcolorbox} 
\usepackage{colortbl}
\usepackage{wrapfig}
\usepackage{enumitem}
\usepackage{stfloats}
\usepackage{booktabs} 
\usepackage{multirow} 
\usepackage{bbm}
\usepackage[dvipsnames]{xcolor}
\definecolor{Gray}{gray}{0.85}

\newcommand{\Lgray}[0]{\rowcolor{gray!10}}

\definecolor{emerald}{rgb}{0.31, 0.78, 0.37}
\definecolor{coralred}{rgb}{1.0, 0.25, 0.25}
\usepackage{tabularx}
\usepackage{subcaption}
\usepackage{float}
\usepackage{placeins}
\def\tsc#1{\csdef{#1}{\textsc{\lowercase{#1}}\xspace}}
\tsc{WGM}
\tsc{QE}

\makeatletter
\providecommand{\frontlastpage}{1}
\providecommand{\mainlastpage}{1}

\AtBeginDocument{%
  \gdef\lastpage{\frontlastpage}%
}

\newcommand{\storefrontlastpage}{%
  \immediate\write\@auxout{%
    \string\gdef\string\frontlastpage{\number\numexpr\value{page}-1\relax}%
  }%
}

\AtEndDocument{%
  \immediate\write\@auxout{%
    \string\gdef\string\mainlastpage{\number\numexpr\value{page}+1\relax}%
  }%
}
\makeatother

\begin{document}
\let\WriteBookmarks\relax
\def\floatpagepagefraction{1}
\def\textpagefraction{.001}

\shorttitle{Geo-R1}    

\shortauthors{Zhang et al.}  

\title [mode = title]{Geo-R1: Improving Few-Shot Geospatial Referring Expression Understanding with Reinforcement Fine-Tuning}

\author[1]{Zilun Zhang}[orcid=0009-0008-5961-5970]
\fnmark[1]
\author[2]{Zian Guan}
\fnmark[1]
\author[3,4]{Tiancheng Zhao}
\cormark[1]

\author[1]{Haozhan Shen}

\author[5]{Yuxiang Cai}

\author[6]{Zhonggen Su}

\author[5]{Yongheng Shang}

\author[7]{Zhaojun Liu}
\cormark[1]

\author[1,5]{Jianwei Yin}
\cormark[1]

\author[8]{Xiang Li}
\ead{xiang92.li@bristol.ac.uk}
\cormark[1]

\affiliation[1]{organization={College of Computer Science and Technology of Zhejiang University},
            city={Hangzhou},
            country={China}}
            
\affiliation[2]{organization={Polytechnic Institute of Zhejiang University},
            city={Hangzhou},
            country={China}}
            
\affiliation[3]{organization={Om AI Research},
            city={Hangzhou},
            country={China}}

\affiliation[4]{organization={Binjiang Research Institute of Zhejiang University},
            city={Hangzhou},
            country={China}}
            
\affiliation[5]{organization={School of Software Engineering of Zhejiang University},
            city={Ningbo},
            country={China}}

\affiliation[6]{organization={School of Mathematical Sciences of Zhejiang University},
            city={Hangzhou},
            country={China}}
            
\affiliation[7]{organization={China Academy of Space Technology},
            city={Beijing},
            country={China}}
            
\affiliation[8]{organization={University of Bristol},
            city={Bristol},
            country={United Kingdom}}

\cortext[cor1]{Corresponding author}

\fntext[eq]{Equal Contribution}

\begin{abstract}
Referring expression understanding in remote sensing poses unique challenges, as it requires reasoning over complex object–context relationships. While supervised fine-tuning (SFT) on multimodal large language models (MLLMs) achieves strong performance with massive labeled datasets, they struggle in data-scarce scenarios, leading to poor generalization. To address this limitation, we propose Geo-R1, a reasoning-centric reinforcement fine-tuning (RFT) paradigm for few-shot geospatial referring. 
Geo-R1 can generate explicit, interpretable reasoning chains that decompose referring expressions, and then leverage these rationales to localize target objects, which provides great interpretability.
We validate Geo-R1 on three carefully designed few-shot geospatial referring benchmarks, where our model consistently and substantially outperforms SFT baselines. It also demonstrates strong cross-dataset generalization, highlighting its robustness. Code and data will be released at \url{https://github.com/Geo-R1/geo-r1}.
\end{abstract}




\begin{keywords}
 \sep Remote Sensing Vision-Language Model 
 \sep Few-shot Learning 
 \sep Referring Expression Task
\end{keywords}


\maketitle


\gdef\lastpage{\mainlastpage}


\section{Introduction}
Vision-language models (VLMs) have become a critical tool for remote sensing (RS) imagery understanding~\citep{li2024vision,weng2025vision}. By coupling natural language with RS imagery, VLMs can drive a wide spectrum of tasks in the RS domain, such as image captioning, \textcolor{black}{visual question answering (VQA), Open-Vocabulary Detection (OVD), Open-Vocabulary Segmentation (OVS)}, referring expression comprehension (REC), referring expression segmentation (RES)~\citep{li2024vision,zhou2024visionlanguagegeofoundationmodelsurvey}. Among these capabilities, REC and RES tasks are especially important: both require the model to resolve free-form linguistic descriptions (e.g., ``a small vehicle is situated at the bottom right adjacent to a large vehicle'') into concrete, spatially localized predictions (bounding boxes or segmentation masks) in high-resolution aerial images. We henceforth use the term \emph{Referring Expression Understanding} (REU) to denote a unified framework encompassing both REC and RES, where the task is to take an image and a text query as input and output one or more target objects.

Although recent works~\citep{kuckreja2023geochatgroundedlargevisionlanguage,yuan2024rrsisreferringremotesensing,zhou2025geogroundunifiedlargevisionlanguage} have achieved remarkable progress on REU tasks with supervised fine-tuning (SFT), these methods are highly dependent on large-scale training labels. High-quality REU supervision demands not only image-level labels but also precise language–region alignment at the object and region levels. Creating such associations in overhead imagery requires expertise and careful tooling: annotators must parse complex scene layouts, disambiguate visually similar man-made structures, and write unambiguous referring expressions before drawing spatially accurate boxes or masks. Compared with image-level labels, these fine-grained annotations are orders of magnitude more labor-intensive. For example, VRSBench~\citep{li2024vrsbenchversatilevisionlanguagebenchmark} 
costs 1,004 labor hours for label verification only.

This reality makes few-shot learning (e.g., only 10 samples are provided for each category) in REU valuable. Previous works, such as RS-CLIP\citep{li2023rsclip} and RemoteCLIP~\citep{liu2024remoteclipvisionlanguagefoundation} have demonstrated that fine-tuning CLIP~\citep{radford2021learningtransferablevisualmodels} on a few samples can yield strong results for scene classification. 
However, these advances cannot be directly carried over to REU since region-level grounding is harder than scene-level classification. Moreover, object relations are complex for REU, requiring relational reasoning and disambiguation among visually similar structures. This raises the question: \textit{with only a handful of aligned examples for each category, can a VLM learn to accurately ground language in remote sensing images?}

\begin{figure*}[htbp]
\begin{center}
\includegraphics[width=\linewidth]{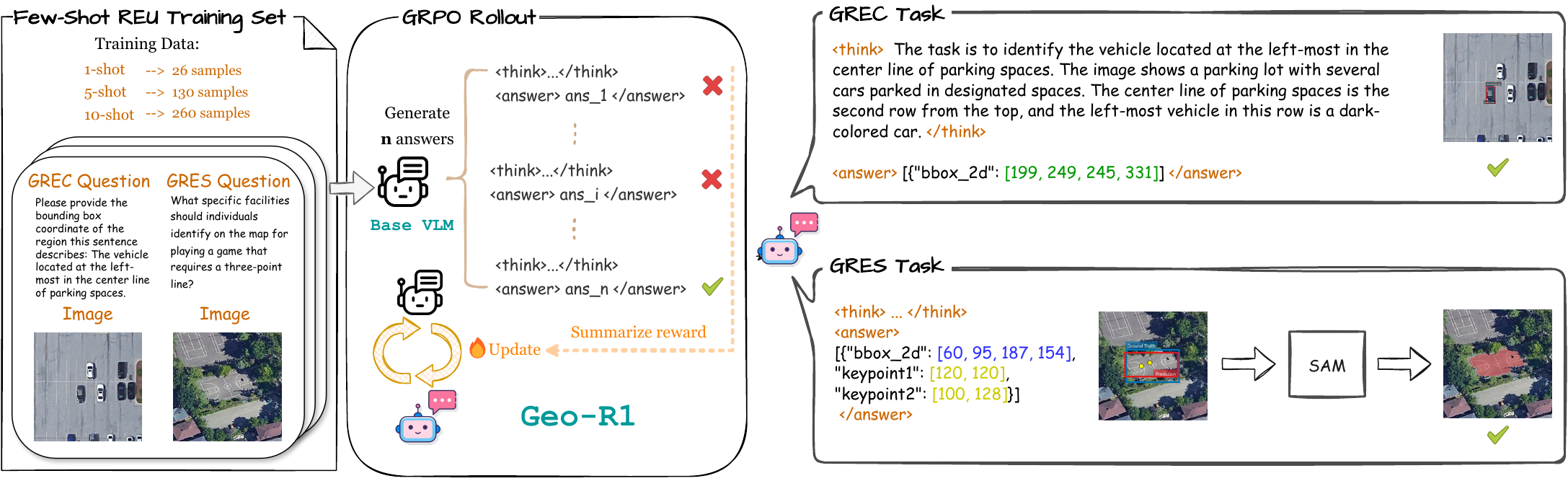} 
\end{center}
\vspace{-10pt}
\caption{Geo-R1 method overview. Geo-R1 is trained on a few labeled samples with reinforcement learning (e.g., GRPO~\citep{shao2024deepseekmathpushinglimitsmathematical}) and can identify target objects (bounding boxes or masks) from an input image and text query while providing the reasoning process.}
\vspace{-10pt}
\label{fig:overview}
\end{figure*}

Driven by the impressive reasoning capabilities of OpenAI o1~\citep{jaech2024openai} and DeepSeek-R1~\citep{deepseekai2025deepseekr1incentivizingreasoningcapability}, reinforcement learning (RL) has become a powerful post-training paradigm for augmenting the reasoning capabilities of large language models (LLMs). RL explicitly encourages intermediate ``thinking'' steps, and forces the model to learn to reason before committing to a prediction. This reasoning-first behavior is particularly well suited to few-shot REU: reasoning steps (e.g., ``My intuition leads me to identify the vehicle sitting in the circular opening near the roadway as the small vehicle.'') serve as a transferable experience that generalizes better across different text-image samples than directly outputting a box/mask from next-token-prediction supervision. \textcolor{black}{Additionally, conventional SFT-based learning paradigm for multimodal LLMs (MLLMs) does not inherently support coordinate regression due to the next-token prediction nature~\citep{jiang2025detectpointprediction}, RL-based framework directly optimizes geometry-aware, task-specific rewards (e.g., box AP, mask-level gIoU), enabling the model to learn spatially grounded behavior that is difficult to obtain through SFT alone. }

In this work, we introduce a reasoning-centric RL post-training method, Geo-R1, which leverages task-specific reward functions to address few-shot REU.
Geo-R1 encourages the model to generate explicit reasoning—intermediate hypotheses that parse the referring expression, identify contextual anchors, and iteratively refine localization—thereby regularizing learning and improving generalization. 
Unlike SFT, which relies on a single teacher-forced trajectory with a differentiable surrogate loss, Geo-R1 explores multiple reasoning chains and proposals, extracting advantages from $N$-way comparisons to provide denser and richer supervision per example, making better use of few-shot samples. 
Moreover, for RES, Geo-R1 directly optimizes a task-aligned \textit{MaskGIoU} reward through the non-differentiable ``BBox + SAM'' pipeline~\citep{ravi2024sam2segmentimages}, enabling end-to-end training for dense prediction—a capability infeasible under SFT. Method overview can be found in Fig.~\ref{fig:overview}.

In our experiments, we observe three consistent advantages from RL over SFT baselines for few-shot REU in remote sensing images. (1) With the same small number of labeled examples, our RFT-based reasoning model substantially outperforms SFT-based models on few-shot REU tasks. (2) In cross-dataset evaluation, our RFT-based model remarkably outperforms SFT counterparts, suggesting the reasoning model has stronger cross-dataset generalization than non-reasoning models. (3) The learned reasoning traces are useful and reasonable, utilizing the spatial and semantic cues that benefit the final localization, which provides a great interpretability.

We further establish three few-shot benchmarks and define a few-shot protocol for REU. In summary, our contributions are listed below:
\begin{itemize}[leftmargin=*]

\item To the best of our knowledge, we are the first to explore Referring Expression Understanding (REU) for aerial image understanding under few-shot settings. To facilitate rigorous and reproducible evaluation, we create VRSBench-FS, EarthReason-FS, and NWPU-FS, establishing standardized protocols for few-shot REU in remote sensing.

\item We define task-aligned rewards and a reasoning-centric RL recipe, including BBoxIoU reward for REC, \textcolor{black}{mAP reward with prediction length penalty for OVD} and a MaskGIoU reward for RES. We introduce the RL-trained reasoning models (Geo-R1) that generate concise grounding rationales for these tasks.

\item Across all three benchmarks, our Geo-R1 models consistently outperform SFT under identical few-shot budgets, while exhibiting stronger generalization across datasets and providing human-auditable reasoning traces that explain successes and failures.

\end{itemize}

\section{Related Work}
\subsection{Reasoning LLMs and VLMs} 
The OpenAI o1~\citep{jaech2024openai} showed that RL improves the reasoning capability of LLMs by learning from feedback on final outcomes. Recently, DeepSeek-R1~\citep{deepseekai2025deepseekr1incentivizingreasoningcapability} demonstrated that rule-based rewards can be used with the GRPO algorithm to teach LLMs advanced reasoning skills. Inspired by the success of RL in LLMs, researchers are now applying the R1 framework to VLMs. R1-OneVision~\citep{yang2025r1} created a step-by-step multimodal reasoning datasets for SFT and RL. Concurrently, R1-V~\citep{chen2025r1v} applied the GRPO algorithm to object counting, achieving the remarkable result of a 3B model outperforming much larger 72B models.
VisualThinker-R1-Zero~\citep{zhou2025r1} applied it directly to base VLMs, observing ``visual aha moments". Other studies refined the training process: Vision-R1~\citep{huang2025vision} first created a multimodal CoT dataset, serving as a cold-start before RL; LMM-R1~\citep{peng2025lmm} used a two-phase strategy, starting with text-only reasoning before fine-tuning on multimodal data. Visual-RFT~\citep{liu2025visualrftvisualreinforcementfinetuning}, VLM-R1~\citep{shen2025vlmr1stablegeneralizabler1style}, and Seg-Zero~\citep{liu2025segzeroreasoningchainguidedsegmentation} explored applying RL to image perception tasks.

\subsection{Few-shot Learning in Remote Sensing}
Few-shot learning (FSL) is crucial in RS, since it effectively addresses the challenge of limited labeled data. Attention-based contrastive learning have been shown to significantly improve classification accuracy in scene classification tasks~\citep{10493055, ZENG2022143}. Prototype-based networks~\citep{li2021few,9435769} and multi-scale feature fusion strategies~\citep{9566599} help models obtain diverse object characteristics, achieving state-of-the-art results on RS object detection benchmarks under few-shot settings. For segmentation, adaptive prototype clustering and mask-guided correlation learning enable precise pixel-level interpretation even with few annotated samples~\citep{jiang2022few,10847788, 10568973, 10613825}. FSL enhances the efficiency and interpretability of RS imagery analysis, while also addressing key challenges in generalization and multimodal integration~\citep{9328476, Lee2024}.

\subsection{REC and RES in Remote Sensing}
Referring expression comprehension in remote sensing—often termed remote sensing visual grounding (RSVG), which localizes a target in aerial imagery from a natural-language description. Early progress was established by the RSVG benchmark and the GeoVG model~\citep{rsvg}, and extended by DIOR-RSVG to broaden categories and scene scale~\citep{diorrsvg}. In the MLLM era, GeoChat~\citep{kuckreja2023geochatgroundedlargevisionlanguage} was the first MLLM to handle a wide range of RS vision-language tasks, including RSVG. Later, VRSBench~\citep{li2024vrsbenchversatilevisionlanguagebenchmark} provided a high-quality dataset for RSVG task. RS-specific MLLMs such as EarthGPT~\citep{zhang2024earthgpt}, RSGPT~\citep{hu2025rsgpt}, SkySenseGPT~\citep{luo2024skysensegpt}, VHM~\citep{pang2025vhm},  further unified different vision-language tasks, such as captioning, VG, VQA, and OVD, thus improving RS-specific alignment. For RES, Yuan et al. introduced the RES task for RS and released the RefSegRS dataset~\citep{yuan2024rrsisreferringremotesensing}. Liu et al. later introduced RRSIS-D, enabling pixel-level referring at scale~\citep{liu2024rotatedmultiscaleinteractionnetwork}. Recent works such as GeoGround~\citep{zhou2025geogroundunifiedlargevisionlanguage} and Skysense-O~\citep{Zhu_2025_CVPR} further unified the REC and RES tasks for RS images. Besides, works for OVD~\citep{li2024toward, pan2025locateearthadvancingopenvocabulary}, and OVS~\citep{li2024segearthovtrainingfreeopenvocabularysegmentation, li2025segearthr1geospatialpixelreasoning} can be viewed as a special case of REC and RES (locate multiple objects with template-based description), which support grounding of novel categories. 

\section{Task and Methodology}
This section details the adaptation of the GRPO algorithm from language-only tasks to vision-language tasks. Then, we introduce and formally define the REU task under few-shot settings. Finally, we discuss how to apply GRPO to these tasks with customized task-specific reward functions.

\subsection{GRPO: from LLM to VLM}
\label{sec:grpo_def}

Group Relative Policy Optimization (GRPO)~\citep{shao2024deepseekmathpushinglimitsmathematical} is a reinforcement learning framework that removes the dependence on a value model and instead utilizes rule-based reward functions. The GRPO algorithm begins by sampling $N$ candidate outputs $\{o_1, \ldots, o_N\}$ from the current policy model $\pi_\theta$ for a given query prompt $q$. Each response $o_i$ is then evaluated by a reward function $R(q, o_i)$ to obtain a raw reward score $r_i$.
\textcolor{black}{To measure the relative quality of each response within the sampled group, GRPO standardizes the raw rewards to obtain the advantage value $\hat{A}_i$, as shown in Eq.~\eqref{eq:relative_advantage}:
\begin{equation}
\begin{aligned}
\hat{A_i} = \frac{r_i - \text{mean}\{r_1, r_2, \ldots, r_N\}}{\text{std}\{r_1, r_2, \ldots, r_N\}}
\label{eq:relative_advantage}
\end{aligned}
\end{equation}
where $\hat{A}_i$ denotes the normalized advantage of the response $o_i$ relative to other samples within the group. }




\textcolor{black}{The policy $\pi_\theta$ is updated by maximizing the GRPO objective in Eq.~\eqref{eq:grpoloss}, which encourages the model to generate responses with higher advantage values:
\begin{equation}
\begin{aligned}
\mathcal{J}_\text{GRPO}(\theta)& = \mathbb{E}_{\{o_i\}_{i=1}^N\sim \pi_{\theta_\text{old}}(\cdot\mid q)} 
\Bigg[ \frac{1}{N}\sum_{i=1}^{N} \Bigg( 
\min \Big( c_1 \cdot \hat{A}_{i},  
\ c_2 \cdot \hat{A}_{i} \Big) \\
&\quad - \beta D_{\text{KL}}(\pi_{\theta} || \pi_{\text{ref}}) 
\Bigg) \Bigg],
\label{eq:grpoloss}
\end{aligned}
\end{equation}
where
\begin{equation}
\begin{aligned}
c_{1} &= \frac{\pi_{\theta}(o_{i} \mid q)}{\pi_{\theta_{\text{old}}}(o_{i} \mid q)}, \\
c_{2} &= \text{clip}\!\left(
\frac{\pi_{\theta}(o_{i} \mid q)}{\pi_{\theta_{\text{old}}}(o_{i} \mid q)},
1 - \varepsilon,\ 1 + \varepsilon
\right).
\end{aligned}
\end{equation}
Here, $D_{\text{KL}}(\pi_{\theta} \,\|\, \pi_{\text{ref}})$ denotes the Kullback--Leibler (KL) divergence between the current policy $\pi_\theta$ and a reference policy $\pi_{\text{ref}}$, which regularizes updates to prevent large deviations. The clipping term $c_2$ stabilizes training by constraining the policy update ratio.}

For LLMs on tasks with definitive answers, like mathematical reasoning, the reward can be calculated using a rule-based verifiable reward function. \textcolor{black}{Building on GRPO}, DeepSeek-R1~\citep{deepseekai2025deepseekr1incentivizingreasoningcapability} demonstrates that such rewards enable models to produce both final answers and coherent reasoning traces. This approach has been successfully extended to VLMs by converting visual metrics into tailored reward signals~\citep{shen2025vlmr1stablegeneralizabler1style,liu2025segzeroreasoningchainguidedsegmentation,liu2025visualrftvisualreinforcementfinetuning}. 

\subsection{Few-Shot Referring Expression Understanding Task}
\label{sec:reu}

\textit{We define referring expression understanding as a unified framework for object recognition (detection or segmentation) from referring expressions.}
\textcolor{black}{Given an image $I$ and a textual query $q$, a vision-language model (VLM) $\mathcal{F}$ predicts one or more target objects as
\begin{equation}
    \{O_1, \ldots, O_N\} = \mathcal{F}(I, q),
    \label{eq:reu}
\end{equation}
where each $O_i$ denotes a predicted object parsed from the VLM's textual output, and $N$ is the number of parsed objects.}

We define REC, Visual Grounding (VG)~\citep{7410660}, and Open-Vocabulary Detection (OVD) as instances of \textit{Generalized REC (GREC)}, where each referred object $O_i$ is represented by a bounding box. Likewise, we define RES and Open-Vocabulary Segmentation (OVS)~\citep{wu2023open} as instances of \textit{Generalized RES (GRES)}, where each object $O_i$ is represented by an instance mask.

In this work, we focus on three representative REU tasks: (i) REC, which targets single-object detection from complex reasoning queries; (ii) OVD, which addresses multi-object detection from class-based queries; and (iii) GRES, which requires multi-object segmentation from complex reasoning queries. All tasks are studied under few-shot settings. In our formulation, each shot label refers to a annotated bounding box or mask. Specifically, in the GREC setup, one ``shot'' is defined as an image–query–box triplet, while in GRES, one ``shot'' corresponds to an image–query–mask triplet. Importantly, a ground-truth mask may include multiple valid instances for a single query~\citep{li2025segearthr1geospatialpixelreasoning}. Among these tasks, GRES is the most challenging, as it requires the model to generate accurate segmentation masks for (multiple) objects described by natural-language queries in aerial images~\citep{yuan2024rrsisreferringremotesensing}.

The few-shot setting substantially increases task difficulty by requiring models to generalize from only a handful of labeled examples, in contrast to large-scale datasets such as VRSBench~\citep{li2024vrsbenchversatilevisionlanguagebenchmark} (36k training examples), and DIOR-RSVG (27k)~\citep{diorrsvg}. Few-shot REU is particularly challenging due to: (1) \emph{visual diversity}, arising from large variations in object size, orientation, appearance, and inter-object relationships; and (2) \emph{description diversity}, as natural language queries may vary in structure, vocabulary, abstraction level, and reasoning complexity. These factors jointly make few-shot REU a more realistic yet significantly harder problem compared to conventional large-scale training scenarios.

\subsection{Reward Design}
\label{sec:reward_design}
Following DeepSeek-R1, the reward function of Geo-R1 includes a task-agnostic format reward and a task-specific metrics reward. The format reward is applied uniformly across all tasks, whereas the metric reward is selected according to the requirements of each specific task.
\textcolor{black}{Our final reward is a combination of the two components:
\begin{equation}
R_{\text{final}} = \lambda_{\text{format}} \, R_{\text{format}} + \lambda_{\text{metrics}} \, R_{\text{metrics}} ,
\end{equation}
where $\lambda_{\text{format}}$ and $\lambda_{\text{metrics}}$ are scalar weights}

\subsubsection{Format Reward}
\textcolor{black}{To ensure reliable parsing, the model’s output must follow a well-defined structure. 
Specifically, the output must be wrapped in the reasoning tags \texttt{<think>\dots</think>} and the answer tags \texttt{<answer>\dots</answer>}. 
Accordingly, we define a binary format reward as:
\begin{align}
R_{\text{format}}(q, o) =
\begin{cases}
1, & \text{if output follows the format}, \\
0, & \text{otherwise}.
\end{cases}
\end{align}
}

\subsubsection{Metrics Reward} 
\noindent\textbf{GREC}. 
For the REC task, the VLM predicts a single bounding box, i.e., $\text{b}_{\text{pred}} = \mathcal{F}\big(I, q)$. An IoU reward can be calculated by comparing $\text{b}_{\text{pred}}$ with the ground-truth box $\text{b}_{\text{gt}}$.
For the OVD task, the VLM predicts is a set of box–label pairs, i.e., $\mathbb{B}_\text{pred} = \{(\text{b}_\text{pred}^i, \text{c}_\text{pred}^i)\}_{i=1}^{N}$, where $\text{b}_\text{pred}^i$ denotes predicted bounding box, $\text{c}_\text{pred}^i$ denotes category label. We then calculate reward as mAP\footnote{We set the confidence score of all predicted bounding boxes to 1.} between $\mathbb{B}_\text{pred}$ and corresponding ground truth $\mathbb{B}_\text{gt}$, along with a penalty coefficient for overlength predictions. \textcolor{black}{Putting these together, the metrics reward for GREC is defined as:}
\begin{align}
R_{\text{metrics}}(q, o) =
\begin{cases}
\text{IoU}(\text{b}_\text{pred}, \text{b}_\text{gt}) \\
\min(\,1, \sqrt{\frac{N_{gt}}{N}}) \cdot \text{mAP}(\,\mathbb{B}_\text{pred}, \mathbb{B}_\text{gt})
\end{cases}
\end{align}
where $N_{\text{gt}}$ denotes the number of ground-truth objects and $N$ is the number of predicted objects. The upper line is the metrics reward for REC task, and the lower line is for OVD task. Both of them belong to GREC tasks.


\noindent\textbf{GRES}. For GRES task, the VLM model is prompted to output a set of box–point pairs, $\mathbb{B}_\text{pred} = \{(b_\text{pred}^i, p_\text{pred}^i)\}_{i=1}^{N}$, where $b_\text{pred}^i$ denotes a predicted bounding box and $p_\text{pred}^i$ denotes the associated keypoints. These predictions are then provided as prompts to a frozen SAM to generate final instance masks $\mathbb{M}_\text{pred}$. Each predicted instance mask is trimmed to ensure its boundary does not exceed that of the corresponding bounding box. Finally, all instance masks are combined by taking their union to form a single predicted segmentation. Given ground truth instance masks $\mathbb{M}_\text{gt}$, the metrics reward for GRES task is defined as:
\begin{align}
R_{\text{metrics}}(q, o) = \text{MaskGIoU}(\mathbb{M}_\text{pred}, \mathbb{M}_\text{gt}).
\end{align} 
We follow LISA~\citep{lai2024lisareasoningsegmentationlarge} to calculate MaskGIoU.

\section{Main Experiment}
\subsection{Experiment Setup}
\label{sec:setup}
\noindent \textbf{Datasets.} Unlike conventional few-shot learning (e.g., Prototypical Networks~\citep{snell2017prototypical} and TFA~\citep{wang2020frustratingly}), we do not partition the dataset into base and novel classes. Instead, we treat all classes as novel and provide only a few labeled examples per class. We construct instruction-following few-shot datasets for the GREC and GRES tasks by deriving them from the training sets of three widely used remote sensing benchmarks: VRSBench~\citep{li2024vrsbenchversatilevisionlanguagebenchmark}, NWPU VHR-10~\citep{CHENG2014119}, and EarthReason~\citep{li2025segearthr1geospatialpixelreasoning}. Configurations and statistics are summarized in Table~\ref{tab:fs_datasets}. \textit{The term ``shot'' defines the number of samples per object category}. For the OVD task, we select four classes on which the baseline model (Qwen2.5-VL-3B) demonstrated decent performance. We select all categories from the training set for other tasks. The low-shot dataset is a subset of the high-shot dataset. To evaluate cross-dataset generalization, we further evaluate zero-shot performance on DIOR-RSVG~\citep{diorrsvg} and RRSIS-D~\citep{yuan2024rrsisreferringremotesensing} datasets.

\begin{table*}[htbp]
    \centering
    \caption{Overview of our Few-Shot Referring Expression Understanding Datasets.}
    \label{tab:fs_datasets}
    \begin{tabular}{c|c|c|c|c|c|c|c}
    \toprule
    \textbf{Dataset Name} & \textbf{Source Dataset} & \textbf{Task} & \textbf{\# Categories} & \textbf{\# Shots} & \textbf{\# VQAs} & \textbf{\# Images} & \textbf{Shot Definition} \\
    \midrule
    \addlinespace 
    VRSBench-FS    & VRSBench       & REC  & 26 & \{10, 5, 1\} & 260   & 254 & image-query-box  \\
    NWPU-FS        & NWPU VHR-10    & OVD & 4  & \{10, 5\} &    25      & 25 &            image-query-box                      \\
    EarthReason-FS & EarthReason    & GRES     & 24 & \{10, 5, 1\} & 240 & 240 & image-query-mask                \\
    \bottomrule
    \end{tabular}
\end{table*}

\noindent \textbf{Model and Training Details.} We adopt Qwen2.5-VL-3B-Instruct~\citep{bai2025qwen25vltechnicalreport} as base model. Our implementation is built on the VLM-R1\footnote{https://github.com/om-ai-lab/VLM-R1} and Easy-R1\footnote{https://github.com/hiyouga/EasyR1} codebase. Unless otherwise specified, we strictly inherit the default hyperparameters without manual tuning. We set the same batch size for different post-training paradigm. We train all compared models for 30 epochs, with early stopping when the reward converged. All experiments are conducted on 8 $\times$ H100 GPUs, and a full training run takes approximately 10 to 20 hours. Prompt templates are shown in Appendix~\ref{sec:prompt_template}. We apply thinking prompts for RL-based Paradigms. We adopt GRPO as our primary RL-based post-training paradigm. For SFT-based post-training, we perform visual instruction tuning with standard next token prediction (NTP) loss, implemented via LLaMA-Factory~\citep{zheng2024llamafactory}. 

\subsection{Few-shot Generalized Referring Expression Comprehension - REC}
\label{sec:recexp}
\noindent \textbf{Task Evaluation.}  Performance on the REC subtask is measured by Acc@$\tau$ (a prediction is correct if its box IoU with the ground truth exceeds $\tau$) in the test set of VRSBench. We report metrics for Acc@0.5 and Acc@0.7. The experiments are conducted in 1-shot, 5-shot, and 10-shot configurations, with ``Unique," ``Non-Unique," and overall results reported. This evaluation compares the SFT method against two RL-based approaches, GRPO~\citep{shao2024deepseekmathpushinglimitsmathematical} and DAPO~\citep{yu2025dapoopensourcellmreinforcement}. We highlight the performance gap in red.

\begin{table*}[b!]
    \caption{Performance on VRSBench for the REC task. We report grounding accuracy at IoU thresholds of 0.5 and 0.7. Unique and Non-Unique indicate whether a referred object is the only instance of its category in the image or not.}
    \label{tab:vg}
    \resizebox{2.0\columnwidth}{!}{%
    \begin{tabular}{l|c|ll|ll|ll}
        \toprule
        \multirow{2}{*}{\textbf{Method}} & \multirow{2}{*}{\textbf{Base LLM}}
        & \multicolumn{2}{c|}{\textbf{Unique}}
        & \multicolumn{2}{c|}{\textbf{Non-Unique}}
        & \multicolumn{2}{c}{\textbf{Overall}} \\
        \cline{3-8}
         & & Acc@0.5 & Acc@0.7 & Acc@0.5 & Acc@0.7 & Acc@0.5 & Acc@0.7 \\
        \midrule
        \multicolumn{7}{l}{\textbf{Full Amount Fine-tune (36,313 samples)  }}\\
        \midrule
        LLaVA-1.5~\citep{liu2024improvedbaselinesvisualinstruction}  & Vicuna1.5-7B & 51.10 & 16.40 & 34.80 & 11.50 & 41.60 & 13.60\\
        Mini-Gemini~\citep{li2024minigeminiminingpotentialmultimodality} & Gemma-7B & 41.10 & 9.60 & 22.30 & 4.90 & 30.10 & 6.80\\
        MiniGPT-v2~\citep{chen2023minigptv2largelanguagemodel} & Vicuna1.5-7B & 40.70 & 18.90 & 32.40 & 15.20 & 35.80 & 16.80\\
        GeoChat~\citep{kuckreja2023geochatgroundedlargevisionlanguage} & Vicuna1.5-7B & 57.40 & 22.60 & 44.50 & 18.00 & 49.80 & 19.90 \\
        Qwen2.5-VL~\citep{bai2025qwen25vltechnicalreport} & Qwen2.5-3B & 66.54 & 36.77 & 60.32 & 36.30 & 62.91& 36.50\\
        \midrule
        \addlinespace[1.2ex] 
        
        \multicolumn{7}{l}{\textbf{Zero-shot Baseline}}\\
        \midrule
        GPT-4V~\citep{GPT4V_SystemCard} &GPT-4& 8.60 & 2.20 & 2.50 & 0.40 & 5.10 & 1.10\\
        \textcolor{black}{Gemini 3.1 Pro} & \textcolor{black}{Gemini 3 Pro} & \textcolor{black}{62.44} & \textcolor{black}{42.72} & \textcolor{black}{53.92} & \textcolor{black}{36.50} & \textcolor{black}{57.55} & \textcolor{black}{39.15} \\
        Qwen2.5-VL w/o thinking & Qwen2.5-3B & 43.10 & 25.10 & 33.46 & 18.01 & 37.48 & 20.97\\
        Qwen2.5-VL w/ thinking & Qwen2.5-3B & 46.18 & 26.90 & 35.22 & 18.87 & 39.79 & 22.22\\
        \midrule
        
        \multicolumn{7}{l}{\textbf{1-shot Fine-tune  (26 samples)}}\\
        \midrule
        Qwen2.5-VL-SFT  & Qwen2.5-3B & 34.32 & 18.87 & 31.62 & 16.35 &  32.75 & 17.40\\
        \Lgray Geo-R1 (GRPO) & Qwen2.5-3B & \textbf{52.17} \tiny \textcolor{black}{(+17.85)} & 31.18 \tiny \textcolor{black}{(+12.31)} & 41.21 \tiny \textcolor{black}{(+9.59)} & 23.04 \tiny \textcolor{black}{(+6.69)} &  45.78 \tiny \textcolor{black}{(+13.03)} & 26.43 \tiny \textcolor{black}{(+9.03)}\\
        \Lgray Geo-R1 (DAPO) & Qwen2.5-3B & 51.72 \tiny \textcolor{black}{(+17.40)} & \textbf{31.68} \tiny \textcolor{black}{(+12.81)} & \textbf{42.13} \tiny \textcolor{black}{(+10.51)} & \textbf{24.50} \tiny \textcolor{black}{(+8.15)} &  \textbf{46.13} \tiny \textcolor{black}{(+13.38)} & \textbf{27.50} \tiny \textcolor{black}{(+10.10)}\\
        \midrule
        \multicolumn{7}{l}{\textbf{5-shot Fine-tune (130 samples)}}\\
        \midrule
        Qwen2.5-VL-SFT  & Qwen2.5-3B & 36.98 & 16.61 & 33.94 & 17.17 &  35.21 & 16.94\\
        \Lgray Geo-R1 (GRPO) & Qwen2.5-3B & 54.11 \tiny \textcolor{black}{(+17.13)} & 31.35 \tiny \textcolor{black}{(+14.74)} & 42.98 \tiny \textcolor{black}{(+9.04)}& 23.98 \tiny \textcolor{black}{(+6.81)}&  47.62 \tiny \textcolor{black}{(+12.41)} & 27.06 \tiny \textcolor{black}{(+10.12)}\\
        \Lgray Geo-R1 (DAPO) & Qwen2.5-3B & \textbf{55.73} \tiny \textcolor{black}{(+18.75)}& \textbf{32.19} \tiny \textcolor{black}{(+15.58)} & \textbf{44.19} \tiny \textcolor{black}{(+10.25)} & \textbf{24.86} \tiny \textcolor{black}{(+7.69)} & \textbf{49.00} \tiny \textcolor{black}{(+13.79)} & \textbf{27.92} \tiny \textcolor{black}{(+10.98)}\\
        \midrule
        \multicolumn{7}{l}{\textbf{10-shot Fine-tune (260 samples)}}\\
        \midrule
        Qwen2.5-VL-SFT  & Qwen2.5-3B & 41.81 & 18.59 & 35.78 & 17.20 &  38.29 & 17.78\\
        \Lgray Geo-R1 (GRPO) & Qwen2.5-3B & 57.27 \tiny \textcolor{black}{(+15.46)} & 35.61 \tiny \textcolor{black}{(+17.02)} & 45.81 \tiny \textcolor{black}{(+10.03)} & 27.03 \tiny \textcolor{black}{(+9.83)} &  50.59 \tiny \textcolor{black}{(+12.30)} & 30.61 \tiny \textcolor{black}{(+12.83)} \\
        \Lgray Geo-R1 (DAPO) & Qwen2.5-3B & \textbf{59.49} \tiny \textcolor{black}{(+17.68)} & \textbf{37.11} \tiny \textcolor{black}{(+18.52)}& \textbf{47.91} \tiny \textcolor{black}{(+12.13)}& \textbf{28.07} \tiny \textcolor{black}{(+10.87)} &  \textbf{52.74} \tiny \textcolor{black}{(+14.45)}& \textbf{31.84} \tiny \textcolor{black}{(+14.06)}\\
        \bottomrule
    \end{tabular}
    }
\end{table*}

\noindent \textbf{Results.}
Table~\ref{tab:vg} compares models trained on the full VRSBench (Full Amount Fine-tune) against few-shot models (1/5/10-shot Fine-tune). The few-shot results include both SFT-based models and our RL-tuned models. Performance data for the full-data baselines (except Qwen2.5-VL) are taken from the original VRSBench paper. The results reveal a clear performance hierarchy: RL-based post-training methods consistently and significantly outperform the SFT approach across all settings and metrics. This advantage is substantial; for example, in the 10-shot overall setting, our GRPO-based model achieves an Acc@0.5 score 12.30\% higher than its SFT counterpart. Remarkably, our 10-shot GRPO model using only 260 samples, 0.71\% data, achieves a score that surpasses all evaluated models (except Qwen2.5-VL) trained on all 36,313 samples.

Within RL-based approaches, DAPO consistently outperforms GRPO across nearly all scenarios, indicating that more effective RL training could further enhance performance in few-shot settings. Moreover, the gains from RL-based methods are more pronounced on the Unique subset than on the Non-Unique subset, suggesting that RL approaches provide a larger boost on simpler tasks that do not require distinguishing between same-category distractors.

\begin{figure*}
\begin{center}
\includegraphics[width=\textwidth]{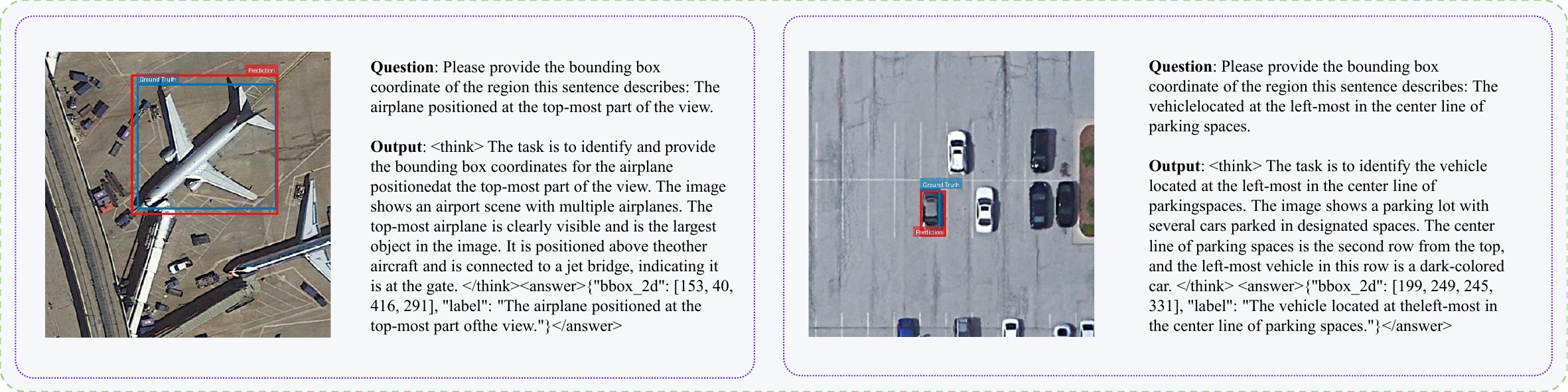}
\end{center}
\vspace{-10pt}
\caption{Geo-R1 inference samples (REC tasks).}
\label{fig:samples2}
\vspace{-10pt}
\end{figure*}

\noindent \textbf{Qualitative Result for REC Task.}
\textcolor{black}{We visualize the model's reasoning process for demonstration. Figure~\ref{fig:samples2} illustrates examples on REC tasks. Geo-R1 is able to provide a reasonable thinking process that leads to an accurate answer. By documenting its steps and observations, the thinking process acts as an internal check, ensuring the model's final bounding box coordinates accurately reflect the object described by the query, leading to more reliable localization.}

\subsection{Few-shot Generalized Referring Expression Comprehension - OVD}

\noindent \textbf{Task Evaluation.} 
For the OVD task, we evaluate performance using the COCO-style mean Average Precision (mAP) in the test set of NWPU VHR-10~\citep{CHENG2014119}. Our evaluation compares the SFT method against our GRPO approach. Experiments are run in 5/10-shot settings. 
Results are reported for the following four categories: airplane (PL), ship (SH), ground track field (GTF), and vehicle (VH). We intentionally exclude the 1-shot setting because training on a single instance would bias the model toward predicting a single instance per image, creating an inconsistency between the training and testing sets. 

\begin{table}[h!]
    \centering
    \caption{Performance on NWPU for the OVD task. We report mAP in COCO style.}
    \label{tab:ovd}
    \resizebox{\columnwidth}{!}{%
    \begin{tabular}{l|c|c|c|c|c}
        \toprule
         & \textbf{PL} & \textbf{SH} & \textbf{GTF} & \textbf{VH} & \textbf{Avg.} \\
        \midrule
        \multicolumn{6}{l}{\textbf{Zero-shot Baseline}}\\
        \midrule
        Qwen2.5-VL w/o thinking & 23.79 & 25.34 & 44.13 & 24.04 & 29.33\\
        Qwen2.5-VL w/ thinking & 25.17 & 21.85 & 57.08 & 23.95 & 32.01\\
        \midrule
        \multicolumn{6}{l}{\textbf{5-shot Fine-tune (20 samples)}}\\
        \midrule
        Qwen2.5-VL-SFT  & 6.32 & 22.33 & 65.48 & 12.36 & 26.62\\
        \Lgray Geo-R1 (GRPO) & \textbf{21.74} & \textbf{25.42} & \textbf{70.23} & \textbf{15.40} & \textbf{33.20}\\
        \Lgray \textcolor{black}{Geo-R1 (DAPO)} & \textcolor{black}{22.15} & \textcolor{black}{25.57} & \textcolor{black}{72.12} & \textcolor{black}{14.70} & \textcolor{black}{33.64}\\
        \midrule
        \multicolumn{6}{l}{\textbf{10-shot Fine-tune (40 samples)}}\\
        \midrule
        Qwen2.5-VL-SFT  & 15.76 & 21.90 & 68.42 & 14.73&30.20\\
        \Lgray Geo-R1 (GRPO) & \textbf{25.76} & \textbf{28.12} & \textbf{69.24} & \textbf{16.57}& \textbf{34.92}\\
        \Lgray \textcolor{black}{Geo-R1 (DAPO)} & \textcolor{black}{29.39} & \textcolor{black}{24.49} & \textcolor{black}{68.20} & \textcolor{black}{17.04} & \textcolor{black}{34.78}\\
        \bottomrule
    \end{tabular}%
    }
\end{table}

\noindent \textbf{Results.} 
Table~\ref{tab:ovd} presents the OVD performance of SFT and GRPO tuned models. A notable observation is that SFT can be detrimental with extremely limited data. In both 10-shot and 5-shot settings, SFT-based models fail to surpass the performance of the zero-shot baseline in three out of four categories (airplane, ship, and vehicle). This suggests that the limited training data lacks intra-class diversity, causing the model to memorize the specific and even spurious features of the few samples rather than the general concept of the class, leading to overfitting, degrading the model's detection capabilities. In contrast, the GRPO-tuned model consistently outperforms the SFT model across all categories and settings, demonstrating that RL is more efficient for learning OVD from a few examples.
More importantly, the advantage of GRPO becomes even more critical in the more challenging low-data setting. The performance gap between GRPO and SFT increases from 4.72 mAP in the 10-shot scenario to 6.58 mAP in the 5-shot scenario. This widening margin highlights GRPO's ability to learn effectively in data-scarce environments where SFT struggles.

\begin{figure*}
\begin{center}
\includegraphics[width=\textwidth]{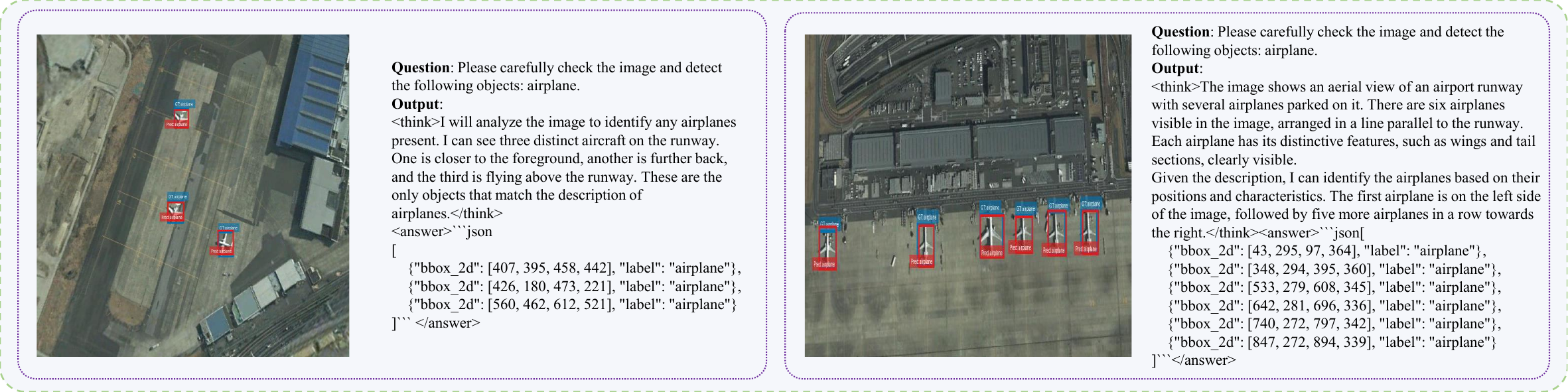}
\end{center}
\vspace{-10pt}
\caption{Geo-R1 inference samples (OVD tasks).}
\label{fig:samples2.5}
\vspace{-10pt}
\end{figure*}

\noindent \textbf{Qualitative Result for OVD Task.}
\textcolor{black}{We visualize the model's reasoning process for demonstration. Figure~\ref{fig:samples2.5} illustrates examples on OVD tasks. The visualizations demonstrate the model's robustness in identifying airplanes across different object densities and scales, successfully detecting sparse, small targets in the first image and differentiating six densely packed aircraft in the second.}

\subsection{Few-shot Generalized Referring Expression Segmentation - GRES}
\noindent \textbf{Task Evaluation.} 
We conduct experiments on the EarthReason~\citep{li2025segearthr1geospatialpixelreasoning} dataset under few-shot setting. Following LISA~\citep{lai2024lisareasoningsegmentationlarge}, performance on the GRES task is measured by the mask-based gIoU, defined by the average of all per-image IoUs. We use this metric because alternatives like cumulative IoU (cIoU) are highly biased toward large-area objects and tend to fluctuate significantly. We report the final gIoU scores on the validation and test sets of the EarthReason dataset. To ensure a fair comparison with SFT-based approaches, we evaluate our method against SegEarth-R1~\citep{li2025segearthr1geospatialpixelreasoning}, trained on the same dataset. SegEarth-R1 serves as a strong SFT baseline, as it employs an auxiliary segmentation decoder to generate pixel-level masks through a differentiable mask loss.

\noindent \textbf{Results.}
In Table~\ref{tab:res}, we demonstrate the effective results of our proposed pipeline and task-specific reward for training reasoning models on GRES task.
\begin{table}[htbp]
    \centering
    \caption{Performance on the EarthReason for the GRES task. We report gIoU.}
    \label{tab:res}
    \resizebox{0.6\columnwidth}{!}{%
    \begin{tabular}{l|cc}
        \toprule
         & \textbf{Val} & \textbf{Test} \\
        \midrule
        \multicolumn{3}{l}{\textbf{Full Amount Fine-tune}}\\
        \midrule
        LISA &  61.04 &  60.88\\
        PixelLM & 57.94 & 60.01\\
        PSALM &  66.61 & 68.30\\
        SegEarth-R1 & 68.60 &  70.75\\
        \midrule
        \multicolumn{3}{l}{\textbf{Zero-shot Baseline}}\\
        \midrule
        Qwen2.5-VL w/ thinking & 19.35 & 32.16 \\
        \midrule
        \multicolumn{3}{l}{\textbf{1-shot Fine-tune  (24 samples)}}\\
        \midrule
        SegEarth-R1  & 42.47 & 43.01\\
        \Lgray Geo-R1 (GRPO)& 50.30 & 51.38 \\
        \Lgray \textcolor{black}{Geo-R1 (DAPO)} & \textcolor{black}{50.09} & \textcolor{black}{51.82}\\
        \midrule
        \multicolumn{3}{l}{\textbf{5-shot Fine-tune  (120 samples)}}\\
        \midrule
        SegEarth-R1  & 45.37 & 45.46 \\
        \Lgray Geo-R1 (GRPO) & 54.73 & 56.01 \\
        \Lgray \textcolor{black}{Geo-R1 (DAPO)} & \textcolor{black}{54.46} & \textcolor{black}{56.24}\\
        \midrule
        \multicolumn{3}{l}{\textbf{10-shot Fine-tune  (240 samples)}}\\
        \midrule
        SegEarth-R1  & 56.40 & 56.60\\
        \Lgray Geo-R1 (GRPO) & 57.78 & 58.27\\
        \Lgray \textcolor{black}{Geo-R1 (DAPO)} & \textcolor{black}{55.56} & \textcolor{black}{58.41}\\
        \bottomrule
    \end{tabular}%
    }
\end{table}
First, we found our GRPO-trained model, i.e., Geo-R1, demonstrates a significant improvement compared to the zero-shot baseline. It achieves a gIoU increase of up to 38.48\% on the validation set (from 19.35\% to 57.78\%) and up to 26.11\% on the test set (from 32.16\% to 58.27\%), showing the success of RL-based post-training paradigm. Then, we observe that the model exhibits remarkable performance with a very small number of samples. With just 240 samples (10-shot), our model demonstrates a comparable performance with PixelLM, which are trained on 900K instances with descriptions. \textit{Using only 240 samples (10-shot), which is roughly 2\% training data, Geo-R1 reaches nearly 83\% of the performance of the SegEarth-R1 model that was trained on the entire training set}.

In a direct comparison, the GRPO pipeline consistently yields superior models to the SFT approach. Geo-R1 outperforms SegEarth-R1 in both the 10-shot and 5-shot settings. Crucially, \textit{this performance gap becomes more significant as the amount of training data decreases}. This trend indicates that RL-based post-training paradigm is a more effective and sample-efficient method for adapting large VLM to this specialized, pixel-level task, especially in data-scarce scenarios.

\begin{figure*}[ht]
\begin{center}
\includegraphics[width=\textwidth]{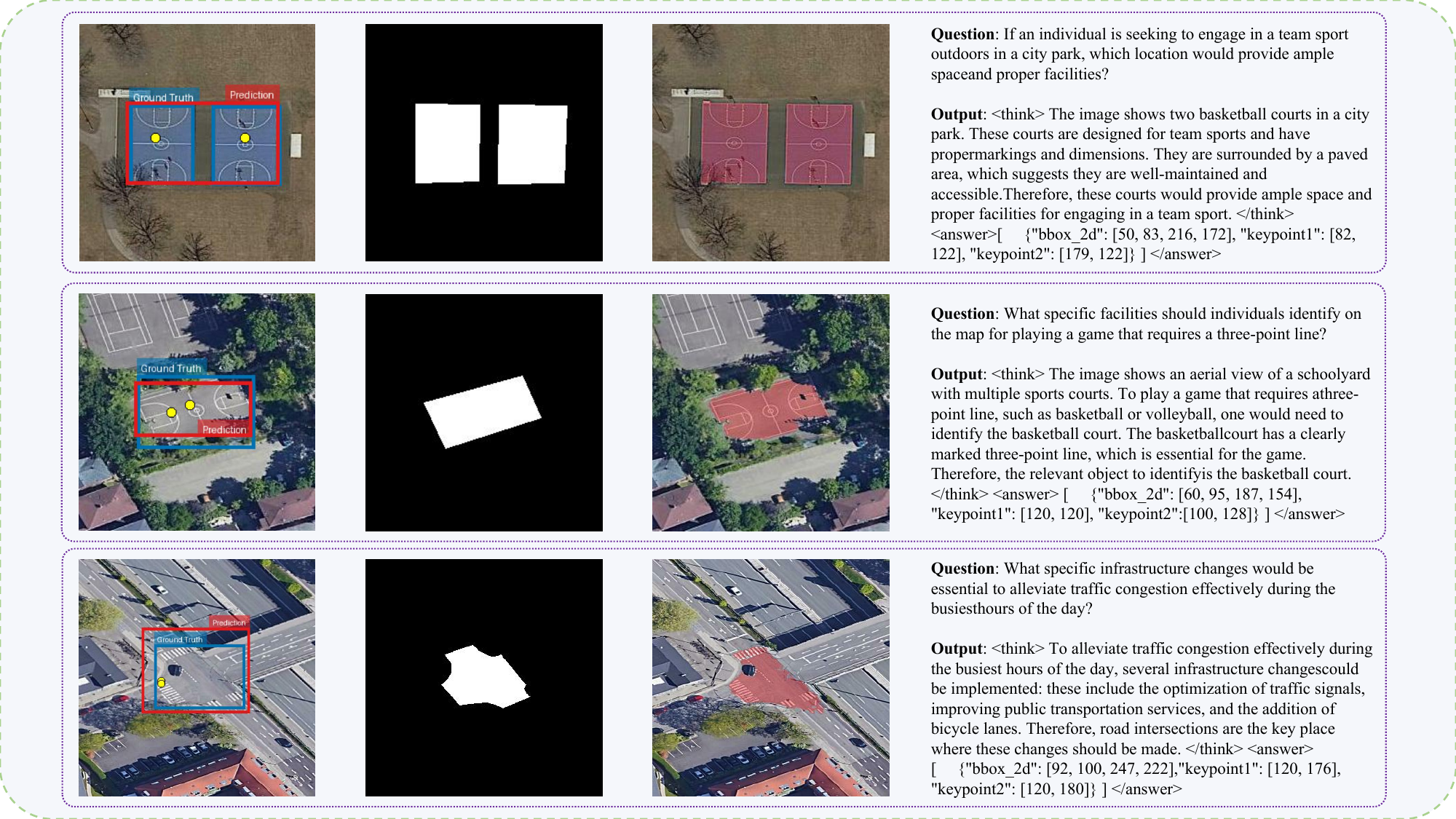}
\end{center}
\vspace{-10pt}
\caption{Geo-R1 inference samples (GRES task).}
\label{fig:samples1}
\vspace{-10pt}
\end{figure*}

\noindent \textbf{Qualitative Result for GRES Task.}
\textcolor{black}{We visualize the model's reasoning process for demonstration. Figure~\ref{fig:samples1} shows some successful case on GRES task. These examples indicate that Geo-R1 is able to perform step-by-step reasoning that leads to accurate segmentation mask.}

\subsection{Cross Dataset Generalization}

\begin{table}[ht]
    \caption{Cross Dataset Evaluation.}
    \resizebox{\columnwidth}{!}{%
    \begin{tabular}{l|c|c|c|c}
        \toprule
         & \multicolumn{2}{c}{VRSBench $\xrightarrow{}$ DIOR-RSVG} & \multicolumn{2}{c}{EarthReason $\xrightarrow{}$ RRSIS-D}\\
        \midrule
         \textbf{\# shot} & \textbf{SegEarth-R1} & \textbf{Geo-R1} & \textbf{SegEarth-R1} & \textbf{Geo-R1}\\
        \midrule
         1-shot & 32.35& 37.27 \tiny \textcolor{black}{(+4.92)} & 18.77 & 32.11 \tiny \textcolor{black}{(+13.34)} \\
        \midrule
         5-shot  & 34.52 & 40.57 \tiny \textcolor{black}{(+6.05)}& 20.29 & 36.41 \tiny \textcolor{black}{(+16.12)} \\
        \midrule
         
         10-shot & 34.86 & 40.38 \tiny \textcolor{black}{(+5.52)} & 24.27 & 37.83 \tiny \textcolor{black}{(+13.56)} \\
        \bottomrule
    \end{tabular}%
    }
    \label{tab:cross}
\end{table}
We further assess the cross-dataset generalization of the SFT and GRPO approaches on the GREC and GRES tasks. For the GREC task, we fine-tune models on the VRSBench dataset with limited supervision (1, 5, and 10-shot) and then evaluate model performance on the DIOR-RSVG target dataset, in a zero-shot manner. As shown in Table~\ref{tab:cross}, GRPO consistently outperforms SFT across all settings, achieving a performance advantage of 4.92\%, 6.05\%, and 5.52\% in the 1-shot, 5-shot, and 10-shot scenarios, respectively. 

Similarly, for the GRES task, models were tuned on the EarthReason dataset (1, 5, and 10-shot) and tested on the RRSIS-D dataset. Here, the GRPO-based model (Geo-R1) demonstrates a remarkable improvement over the SFT-based model (SegEarth-R1) under few-shot setting, achieving a relative improvement up to 80\%. These results highlight GRPO's incredible cross-dataset generalization, indicating superior transferability and robustness of Geo-R1.

\subsection{Geo-R1 for SAR Imagery}
\label{sec:sar}

\textcolor{black}{The effectiveness of RL-based post-training depends strongly on the capability of the base model. In other word, it may require modality-specific preprocessing or stronger modality-adapted encoders for MLLM. At present, we have not found any open-sourced MLLM that is explicitly designed for multispectral, hyperspectral, and SAR imagery and that directly supports REC/RES on these modalities. For example, EarthDial~\citep{Soni_2025_CVPR} is open-sourced and uses multispectral and SAR instruction-tuning data for SFT. However, it only releases a multispectral checkpoint and focuses on classification. A similar limitation holds for CHOICE~\citep{an2025choicebenchmarkingremotesensing}, which mainly targets multiple-choice question answering. SARCLIP~\citep{11231801} and SARVLM~\citep{ma2025sarvlmvisionlanguagefoundation} are CLIP-based models and neither target MLLM-style instruction following nor support REC/RES.
SARChat~\citep{Ma2025SARChatBench2MAM} is a MLLM that supports SAR, but it primarily focuses on detection-style tasks, and its textual data is not directly suitable for complex referring expressions required by REC/RES.}

\textcolor{black}{To provide concrete evidence on SAR, we further construct a SAR few-shot REC benchmark by following the VRSBench annotation protocol and scripts. Specifically, we use GPT-4o to annotate a subset of SARDet-100K \footnote{https://huggingface.co/datasets/omlab/SARDet\_REC6-FS} and release it as an open dataset. The dataset contains six categories: Ship, Aircraft, Car, Bridge, Harbour, and Tank. Its train/test splits are summarized in Table~\ref{tab:sarrec6_dist}. }

\begin{table}[t]
\centering
\caption{Class distribution of SARDet\_REC6-FS (train/test).}
\label{tab:sarrec6_dist}
\setlength{\tabcolsep}{6pt}
\renewcommand{\arraystretch}{1.0}
\begin{tabular}{c|cccccc}
\toprule
\textbf{Class} & \textbf{Ship} & \textbf{Aircraft} & \textbf{Car} & \textbf{Bridge} & \textbf{Harbour} & \textbf{Tank} \\
\midrule
\textbf{Train} & 10 & 10 & 10 & 10 & 10 & 10 \\
\textbf{Test}  & 200 & 200 & 80 & 200 & 200 & 28 \\
\bottomrule
\end{tabular}
\end{table}

\textcolor{black}{Same as in Section~\ref{sec:recexp}, we train on the training split of SARDet\_REC6-FS (6 classes, 10-shot per class) and report Acc@0.5 on the test split. The results are summarized in Table~\ref{tab:sarrec6_rec_results}. Geo-R1 outperforms SFT-based model on few-shot SAR REC task under the same 10-shot budget. The SFT model yields a slight negative effect compared to the zero-shot baseline, while Geo-R1 achieves consistent Acc@0.5 gains across all six categories, indicating that geometry-aligned, on-policy RFT is more sample-efficient and robust than SFT even on SAR imagery.}

\begin{table}
\centering
\caption{Per-category REC results on SARDet\_REC6-FS (test). We report Acc@0.5.}
\label{tab:sarrec6_rec_results}
\footnotesize
\setlength{\tabcolsep}{3pt}
\renewcommand{\arraystretch}{1.0}
\resizebox{\columnwidth}{!}{%
\begin{tabular}{c|c|cccccc}
\toprule
\textbf{Method} & \textbf{Mean} & \textbf{Ship} & \textbf{Aircraft} & \textbf{Car} & \textbf{Bridge} & \textbf{Harbour} & \textbf{Tank} \\
\midrule
Qwen2.5-VL & 15.31 & 29.50 & 8.50  & 3.75 & 3.00 & 27.00 & 0.00 \\
Qwen2.5-VL-SFT       & 14.87 & 28.00 & 7.50  & 3.75 & 3.50 & 27.00 & 0.00 \\
Geo-R1 (GRPO)    & 25.00 & 56.00 & 15.00 & 20.00 & 5.00 & 28.50 & 7.14 \\
\bottomrule
\end{tabular}
}
\end{table}

\section{Discussion}
In this section, we conduct extensive experiments to validate method designs. Unless otherwise specified, experiments are conducted on the VRSBench-FS dataset under the 10-shot setting with GRPO.  

\subsection{Upper bound of Few-shot Learning}
As shown in Fig.~\ref{fig:upperbound}, GRPO clearly outperforms SFT in low-shot settings, although this performance gap narrows as supervision increases. To investigate this trend and determine the upper-bound of Geo-R1, we experimented with additional shot numbers (20, 50, 100, and 200).
Concretely, the margin between GRPO and SFT approaches shrinks from 13.03\% at 1-shot to 0.47\% at 200-shot. This diminishing advantage suggests both approaches approach a common upper bound with more data. Empirically, they converge toward the full-data SFT result of 62.91\%, indicating GRPO’s strong sample efficiency at small shots but similar asymptotic performance as shot count grows.


\begin{center}
\includegraphics[width=\columnwidth]{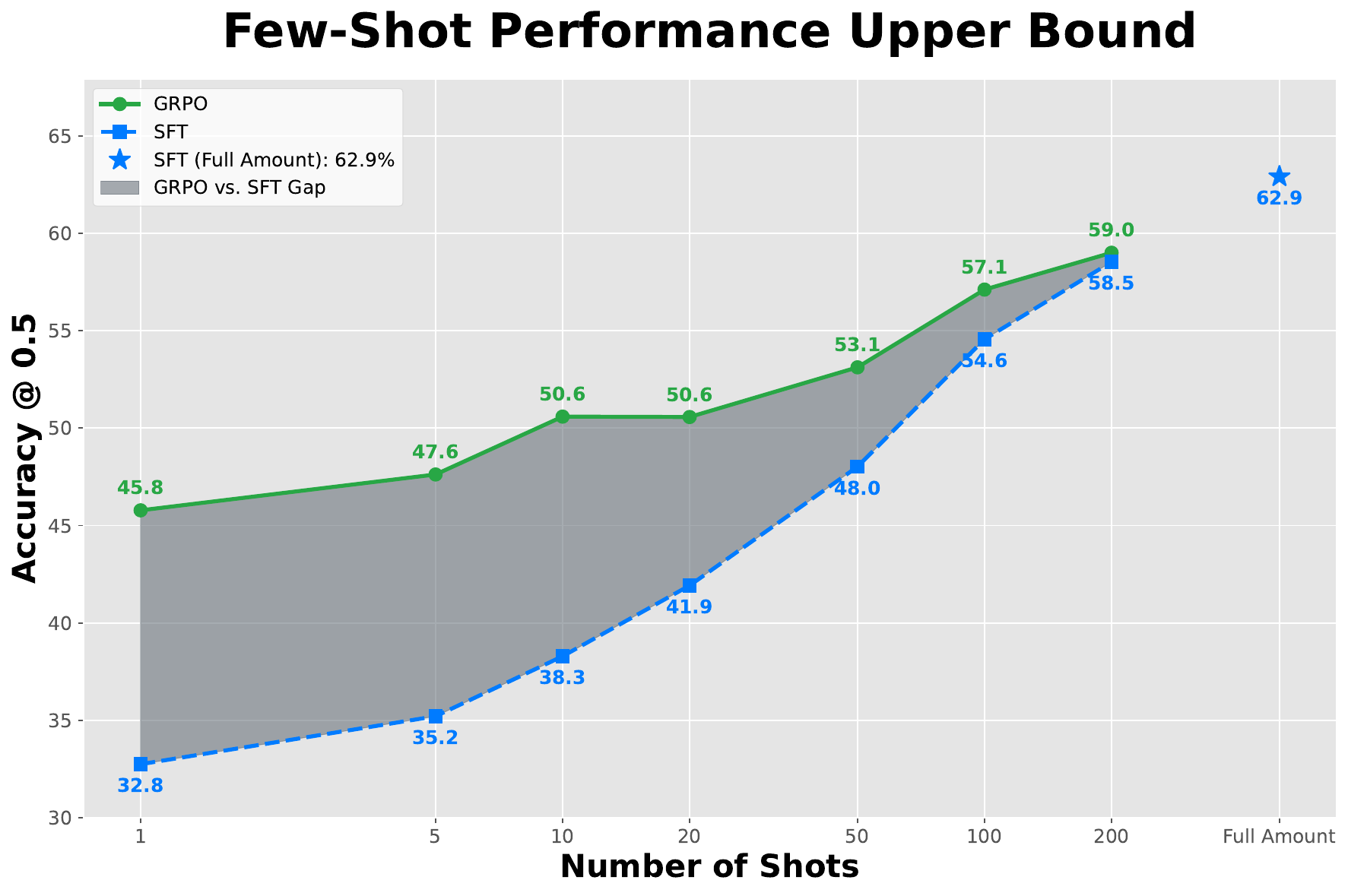}
\captionof{figure}{Few-shot Learning Upper-Bound.}
\label{fig:upperbound}
\end{center}

\subsection{Learning Curve Comparison}
We fine-tune Qwen2.5-VL-3B with both SFT and GRPO on the REC task using same batch size and evaluate checkpoints every 100 steps to sketch the learning curve.
As shown in Fig.~\ref{fig:learningcurve}, GRPO consistently outperforms SFT at every checkpoint, with an average gain of 9.74\%. GRPO improves steadily, peaking around 400 steps, and remains strong until the end, whereas SFT oscillated within 37\%–40\%. GRPO achieves a clearly higher ceiling and stabilizes around 50\%, indicating better training efficiency under few-shot setting.


\begin{center}
\includegraphics[width=\columnwidth]{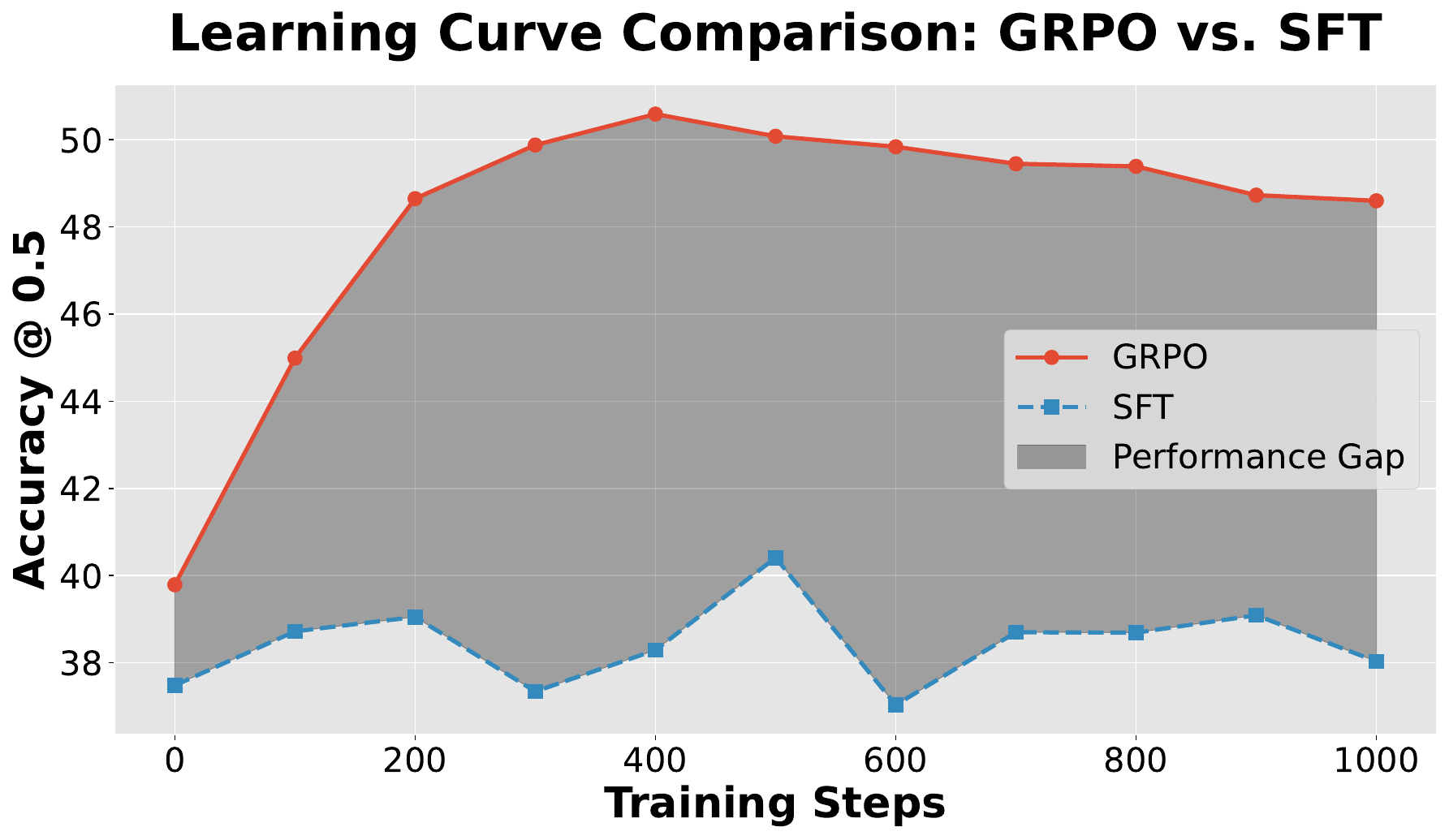}
\captionof{figure}{Learning curves of GRPO vs. SFT on REC.}
\label{fig:learningcurve}
\end{center}

\subsection{Comparison with Other Few-shot Approaches}
\label{sec:gdino}
\textcolor{black}{We add a comparison, Grounding-DINO~\citep{liu2024groundingdinomarryingdino}, for REC task. Specifically, the Grounding-Dino-T is fine-tuned with 10-shot and 5-shot REC data for 30 epochs, using the Open-Grounding-Dino framework\footnote{https://github.com/longzw1997/Open-GroundingDino}. The result can be seen in Table~\ref{tab:gdino}. Geo-R1 consistently outperforms both Grounding-DINO and the SFT baseline across all few-shot settings. Notably, the advantage becomes more pronounced as the amount of training data decreases when compared with Grounding-DINO. In particular, under the 5-shot and 1-shot settings, the performance gap further widens. This result indicates that RL-based post-training strategy for MLLM yields stronger generalization and better adapts to REC than Grounding-DINO when limited supervision is available.}

\begin{table}
    \caption{\textcolor{black}{Comparison with other few-shot approaches.}}
    \color{black}
    \centering
    \begin{tabular}{l|c|c|c}
        \toprule
         & 10-shot & 5-shot & 1-shot \\
        \midrule
        Grounding-DINO & 37.65	&  26.52 & 22.51\\
        Qwen2.5-VL-SFT & 38.29	&  35.21 & 32.75\\
        Geo-R1	& 50.59	& 47.62 & 45.78\\
        \bottomrule
    \end{tabular}%
    \label{tab:gdino}
\end{table}

\subsection{Ablation Study}

\noindent \textbf{Effect of model size.}
We examine how model size influences the performance under different post-training paradigms. As shown in Fig.~\ref{fig:modelsize}, both SFT and GRPO benefit from increased model scales.
However, this trend exhibits clear diminishing marginal returns. For instance, SFT gains 4.31\% when scaling from 3B to 7B but only 2.23\% from 7B to 32B, with a similar slowdown observed for GRPO from 3B to 7B. This suggests that while larger models provide a stronger foundation, simply increasing number of parameters yields limited benefits for the few-shot task. 
This can be attributed to the limited fine-tuning data. With few examples, high-capacity models tend to overfit by simply memorizing the training samples rather than learning generalizable features.
Notably, GRPO's performance decreased on the 32B model, likely due to overfitting on limited data when using more clever model.


\begin{center}
\includegraphics[width=\columnwidth]{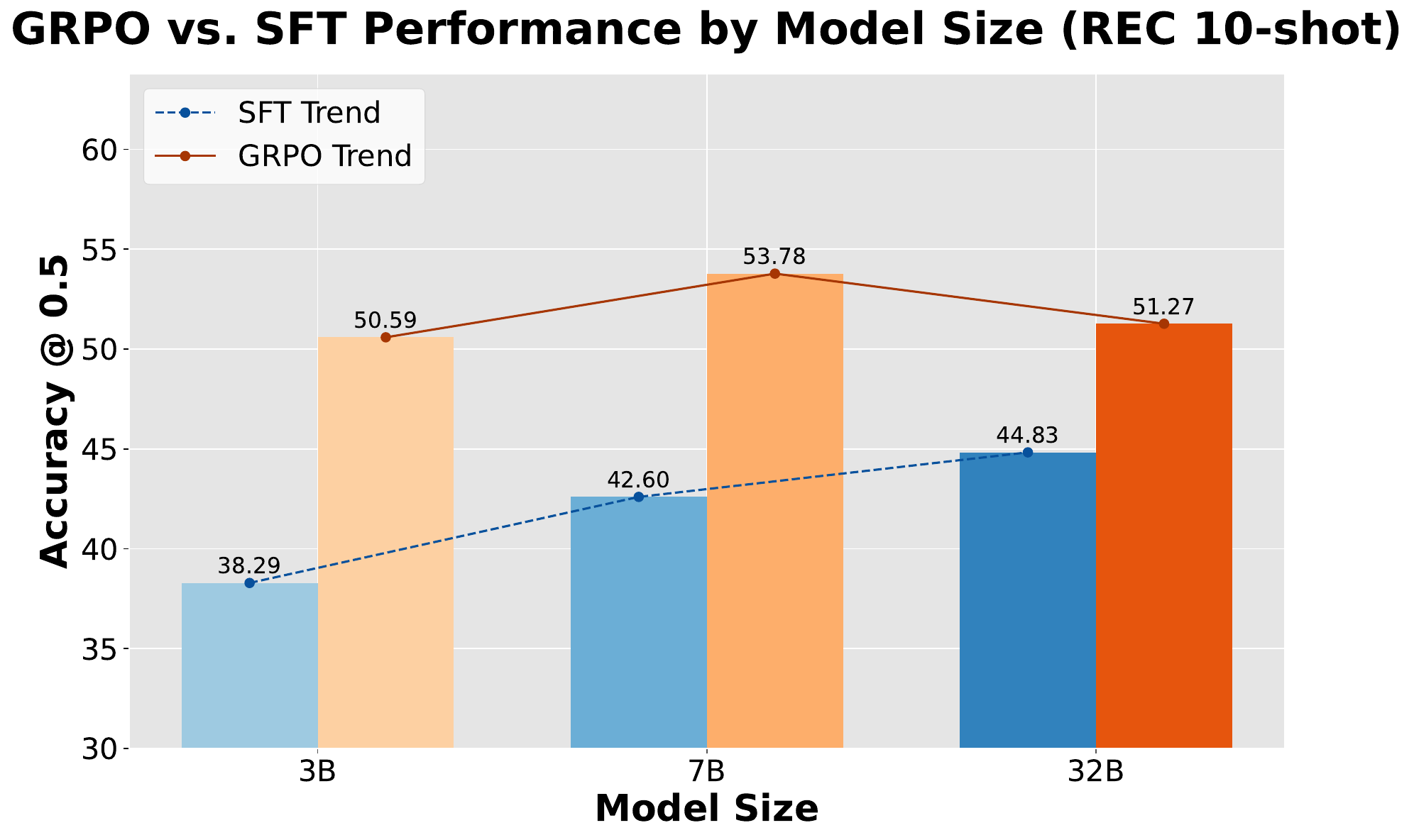}
\captionof{figure}{Few-shot Learning Meets Model Size.}
\label{fig:modelsize}
\end{center}

\noindent \textbf{Effect of Thinking.}
We additionally include results for our RL-based method without chain-of-thought reasoning (i.e., the model predicts the final answer directly). As shown in the Table~\ref{tab:thinking}, Geo-R1 w/o thinking performs slightly worse than Geo-R1 w/ thinking, but still substantially outperforms the SFT counterpart. This indicates that the performance gains primarily stem from the training paradigm (geometry-aware RL) rather than from the reasoning process itself.
\begin{table}[htbp]
    \centering
    \caption{Comparison on models with and without Thinking}
    \begin{tabular}{l c c}
        \toprule
         & REC (Acc@0.5) & OVD (mAP) \\
        \midrule
        SFT & 38.29 & 30.20 \\
        Geo-R1 w/ thinking & 50.59 & 34.92 \\
        Geo-R1 w/o thinking & 48.10 &  32.79\\
        \bottomrule
    \end{tabular}
    \label{tab:thinking}
\end{table}

\noindent \textbf{Effect of KL Divergence.}
We explicitly examine the effect of KL regularization under the 10-shot setting for all three REU tasks, using GRPO. As shown in Table~\ref{tab:kldiver}, the impact of KL regularization is task-dependent.
For REC, adding KL brings a clear improvement, suggesting that KL stabilizes policy updates and helps prevent the model from drifting too far from the SFT initialization when learning fine-grained spatial grounding.
For OVD, the effect is marginal, indicating that KL plays a limited role in this setting.
For GRES, removing KL slightly improves performance, which may indicate that mask-level rewards can require larger policy updates for effective exploration. In such cases, KL may slightly limit exploration for GRES.

\begin{table}[htbp]
    \centering
    \caption{Effect of KL regularization under 10-shot GRPO training.}
    \label{tab:kldiver}
    \setlength{\tabcolsep}{6pt}
    \renewcommand{\arraystretch}{1.15}
    \begin{tabular}{lcc}
        \toprule
        \textbf{Task (metric)} & \textbf{With KL} & \textbf{Without KL} \\
        \midrule
         REC (Acc@0.5) &  50.59	&  44.72\\
         OVD (mAP) &  34.92 &  34.59 \\
         GRES (gIoU) &  58.27	&  59.26\\
        \bottomrule
    \end{tabular}
\end{table}


\noindent \textbf{Effect of OVD Length Penalty.} Table~\ref{tab:overlongreward} shows that without the length penalty reward, Geo-R1 exhibits significantly degraded performance on the OVD task.

\begin{table}[ht]
    \caption{Ablation on Length Penalty of OVD task.}
    \centering
    \begin{tabular}{l|c}
        \toprule
         & OVD (mAP)\\
        \midrule
        Geo-R1 w/ length penalty & 34.92 \\
        Geo-R1 w/o length penalty & 27.52 \\
        \bottomrule
    \end{tabular}%
    \label{tab:overlongreward}
\end{table}

\noindent \textbf{Comparison with Direct Preference Optimization (DPO).} We construct the DPO training dataset and train a DPO model on it. Details can be found in Appendix ~\ref{sec:dpo_construct}. The DPO results are reported in Table~\ref{tab:dpo}.
DPO relies on offline preference pairs. In REC, small numerical deviations in coordinates can cause large IoU drops, making pairwise preferences highly sensitive and sometimes unstable. As a result, DPO can only provide a weaker and noisier learning signal than GRPO/DAPO, which directly optimizes geometry-aware rewards with on-policy self-exploration. Moreover, DPO can fail when the preference data does not cover the model’s dominant error modes. In that case, DPO provides little gradient signal on critical grounding mistakes, so the model is not consistently pushed away from its key failure regions. This problem is amplified under few-shot supervision and can be mismatched to the true error distribution, making DPO far less sample-efficient than GRPO/DAPO.

\begin{table}[ht]
    \centering
    \caption{REC results on VRSBench-FS (10-shot): comparison of self-exploration RFT (GRPO/DAPO), and preference-based RFT (DPO).}
    \label{tab:dpo}
    \begin{tabular}{l|c|c}
        \toprule
        Method & Training approach & Acc@0.5 \\
        \midrule
        Geo-R1 (GRPO)    & Self-exploration-based RFT           & 50.59 \\
        Geo-R1 (DAPO)    & Self-exploration-based RFT           & 52.74 \\
        Geo-R1 (\textbf{DPO})     & Preference-based RFT                 & 37.34 \\
        \bottomrule
    \end{tabular}
\end{table}



\subsection{More Experimental Results}
\label{sec:appendix_exp}
To provide a more comprehensive evaluation beyond the main paper, we include additional experiments and analyses in the apendix. These include:
(i) robustness analysis under multiple random seeds;
(ii) quantitative assessment of generated reasoning chains;
(iii) detailed failure case analysis; and
(iv) training stability analysis across tasks.
The results consistently corroborate our main findings and further demonstrate the robustness and effectiveness of Geo-R1.

\subsection{Limitation}
\textcolor{black}{
In our GRES pipeline, the MLLM does not generate pixel-wise masks directly. Instead, it predicts structured prompts (e.g., bounding boxes and keypoints), and the final segmentation mask is produced by a frozen segmentation model (SAM2). RL fine-tuning is performed by computing a mask-level reward on the final output mask and backpropagating the resulting policy gradient updates to the MLLM. We optimize the prompt generation policy using mask-level supervision, rather than updating the segmenter itself. This decoupled design also has limitations: the performance depends on the segmentation model’s prior to remote-sensing imagery and may be bounded if the segmentation model is not good at remote sensing imagery. On the other hand, this setup is flexible. The segmentation model can be replaced with stronger or domain-adapted models (e.g., SAM3~\citep{carion2025sam3segmentconcepts}, RemoteSAM~\citep{yao2025remotesamsegmentearthobservation}), and leveraging a highly generalizable prompt-based segmentation model enables the system to achieve strong GRES performance with only a small number of training samples by primarily learning effective prompts.
}

\section{Conclusion}
In this work, we define a generic task, Referring Expression Understanding that aims to recognize objects (either detection/segmentation) from referring expressions. We then compare RL-based (GRPO) and SFT-based post-training paradigms on few-shot REC, OVD, and GRES tasks within the RS domain. Our results show that our GRPO-trained model, Geo-R1, consistently outperforms standard SFT-tuned models across these tasks. The performance gains are particularly large in low-shot regimes, and the model exhibits significantly stronger cross-dataset generalization.

While our study demonstrates the effectiveness of reinforcement learning for few-shot referring expression understanding, several avenues remain. Our evaluation is limited to high-resolution aerial imagery; extending Geo-R1 to multispectral (e.g., Sentinel-2) and SAR data would further test its robustness. Beyond the three REU tasks studied (REC, RES, OVD), future work could explore broader grounding tasks, e.g., OVS. Finally, scaling to larger shots, refining reward functions, and designing powerful RL training recipes remain promising directions.

\section{Acknowledgements}
This work is supported by the National Key Research and Development Program of China (Grant 2024YFB3908402), Hangzhou High-Tech Zone (Binjiang) platform-based open-competition project (2025JBGS-PT003), and the National Key Research and Development Program of China (Grant 2024YFF1400900).

\section{Declaration of AI-assisted technologies in the manuscript preparation process.}

During the preparation of this work the authors used ChatGPT and Gemini in order to (i) edit and polish grammar and phrasing; (ii) use ``DeepResearch'' to help retrieve and cluster related literature (with all citations verified by the authors). (iii) Initialized a draft for CoT evaluation prompt. After using this tool/service, the authors reviewed and edited the content as needed and take full responsibility for the content of the published article.





\clearpage
\newpage
\twocolumn
\bibliographystyle{cas-model2-names}
\bibliography{cas-refs}



\newpage
\appendix
\onecolumn
\begin{center}
    \LARGE \textbf{Supplementary of Geo-R1: Improving Few-Shot Geospatial Referring Expression Understanding with Reinforcement Fine-Tuning}
\end{center}
\vspace{1em} 
\section{Geo-R1 Prompt Template}
\label{sec:prompt_template}
We largely follow the VLM-R1 prompt templates for REC, OVD, and extend the same interface to the GRES setting. We append the thinking template at the end for all task prompts.

\tcbset{
  colback=gray!5,
  colframe=gray!40!black,
  fonttitle=\bfseries,
  breakable,
}

\begin{tcolorbox}[
    title=Prompt Template of REC  
]
Please provide the bounding box coordinates of the region this sentence describes: \{\texttt{query}\}.
\end{tcolorbox}

\begin{tcolorbox}[
    title=Prompt Template of OVD, 
    breakable, 
    enhanced jigsaw
]
Please carefully check the image and detect the following objects: \{\texttt{target list}\}.
Output each detected target's bbox coordinates in JSON format. The format of the bbox coordinates is:

\begin{verbatim}
```json
[
{
    "bbox_2d": [x1, y1, x2, y2], 
    "label": "category name"
},
{
    "bbox_2d": [x1, y1, x2, y2], 
    "label": "category name"
}
]
```
\end{verbatim}
If there are no such targets in the image, simply respond with None.
\end{tcolorbox}

\newpage
\begin{tcolorbox}[
    title=Prompt Template of GRES, 
    breakable, 
    enhanced jigsaw
]
Please carefully check the image and answer: \{\texttt{query}\}. Based on your answer, detect all relevant objects in the image. Output each detected target's bbox coordinates in JSON format. The format of the bbox coordinates is:
\begin{verbatim}
```json
[
{
    "bbox_2d": [x1, y1, x2, y2], 
    "keypoint1": [x3, y3], 
    "keypoint2": [x4, y4]
},
{
    "bbox_2d": [x1, y1, x2, y2],
    "keypoint1": [x3, y3],
    "keypoint2": [x4, y4]
}
]
```
\end{verbatim}
\end{tcolorbox}

\begin{tcolorbox}[
    title=Thinking Template  
]
\texttt{\{problem\}} Output the thinking process in $<$think$>$ $<$/think$>$ and final answer in $<$answer$>$ $<$/answer$>$ tags.
\end{tcolorbox}

\section{Evaluation Prompt}
\label{sec:eva_prompt_template}
As mentioned in section~\ref{sec:assess}, we carefully designed prompts to assess both the correctness and usefulness of the reasoning toward the final prediction, assigning score on a 1–10 scale. The prompt is listed below.

\tcbset{
  colback=gray!5,
  colframe=gray!40!black,
  fonttitle=\bfseries,
  breakable,
}

\begin{tcolorbox}[title={CoT Quality Evaluation Prompt}]

\textbf{You are a strict evaluator} for a vision--language model that performs
\textbf{referring expression comprehension (visual grounding)} on remote sensing images.

\medskip

For each sample, you will be given:
\begin{itemize}
  \item \texttt{image}: a remote sensing / aerial image.
  \item \texttt{question}: a natural--language referring expression describing a target region.
  \item \texttt{ground\_truth}: the correct bounding box of the referred region:
        \texttt{[x\_min, y\_min, x\_max, y\_max]}.
  \item \texttt{model\_output}: the model’s full output, which contains:
  \begin{itemize}
    \item a reasoning \textbf{Chain-of-Thought} between \texttt{<think> ... </think>},
    \item a final JSON-style prediction between \texttt{<answer> ... </answer>}
          with a predicted bbox \texttt{bbox\_2d}.
  \end{itemize}
  \item \texttt{extracted\_answer}: the predicted bbox
        \texttt{[x\_min, y\_min, x\_max, y\_max]} parsed from the final answer.
  \item \texttt{correct}: whether the final prediction is counted as correct (1 or 0).
\end{itemize}

Your job: \textbf{only evaluate the reasoning text inside \texttt{<think>...</think>}}
and output \textbf{a single overall quality score from 1 to 10} (higher is better).

You should internally consider two aspects:
\begin{enumerate}
  \item \textbf{Correctness} of the reasoning.
  \item \textbf{Usefulness} of the reasoning.
\end{enumerate}
But your final output must be just \textbf{one combined score} that reflects both.

Do \textbf{not} change the model’s prediction. Do \textbf{not} generate a new bounding box.

\medskip\hrule\medskip

\textbf{1.\ Criterion to Consider When Scoring}

\textbf{1.1.\ Correctness}

Correctness measures how \textbf{factually accurate and logically sound} the reasoning is,
relative to:
\begin{itemize}
  \item the image content,
  \item the question / referring expression,
  \item the ground-truth bounding box and the predicted bounding box.
\end{itemize}

Key points:
\begin{enumerate}
  \item \textbf{Visual faithfulness}
  \begin{itemize}
    \item The reasoning should accurately describe what is visible (object categories,
          spatial relations, comparative relations, etc.).
    \item Penalize:
    \begin{itemize}
      \item referring to objects that do not exist in the image,
      \item clearly wrong locations (e.g., calling a top--left object ``bottom--right''),
      \item confusing object types (e.g., calling a tennis court a baseball field).
    \end{itemize}
  \end{itemize}

  \item \textbf{Consistency with ground-truth and prediction}
  \begin{itemize}
    \item Use \texttt{ground\_truth} and \texttt{extracted\_answer} to understand
          which region is correct and how close the prediction is.
    \item If the predicted bbox is near the ground truth and the reasoning clearly
          supports this localization, correctness should be \textbf{high}, as long
          as there are no serious hallucinations.
    \item If the predicted bbox is far from ground truth and the reasoning is based
          on the wrong object/region, correctness should be \textbf{low}, even if the
          text is fluent.
  \end{itemize}

  \item \textbf{Logical consistency}
  \begin{itemize}
    \item The reasoning should be internally coherent:
    \begin{itemize}
      \item no self-contradiction (e.g., first says ``top--left'' then ``bottom--right''
            for the same object),
      \item no obvious contradiction with the predicted bbox (e.g., reasoning says
            the object is ``top--left'' but the predicted box is in bottom--right).
    \end{itemize}
  \end{itemize}
\end{enumerate}

\noindent\textbf{Rule:} If correctness is extremely low (e.g., the reasoning clearly
focuses on a completely wrong object or is mostly hallucinated), the final overall
score must also be low (around 1--3), no matter how nicely written it is.

\medskip

\textbf{1.2.\ Usefulness}

Usefulness measures how \textbf{helpful and informative} the reasoning is for
explaining \textbf{why} the model chose the final bounding box.

Key points:
\begin{enumerate}
  \item \textbf{Task-specific grounding}
  \begin{itemize}
    \item A useful CoT clearly connects the question to the image:
    \begin{itemize}
      \item identifies candidate objects that match the category (ships, vehicles,
            fields, buildings, etc.),
      \item compares positions (left-most, bottom-most, near center, near boundary, etc.),
      \item uses context (e.g., ``among the ships'', ``in the bottom-left corner'',
            ``between two runways'', ``inside the stadium'', ``on the water, not on land'').
    \end{itemize}
    \item Penalize:
    \begin{itemize}
      \item CoT that just paraphrases the question without adding visual evidence,
      \item very generic lines (``I look at the image and find the object described'')
            with no actual reasoning.
    \end{itemize}
  \end{itemize}

  \item \textbf{Conciseness and relevance}
  \begin{itemize}
    \item Prefer reasoning that is focused on the grounding task.
    \item Penalize:
    \begin{itemize}
      \item very long but repetitive reasoning that adds no extra information,
      \item completely off-topic storytelling.
    \end{itemize}
  \end{itemize}
\end{enumerate}

\medskip\hrule\medskip

\textbf{2.\ How to Combine into a Single Score (1--10)}

You must convert your internal judgment about correctness and usefulness into
a \textbf{single integer score from 1 to 10}.

Use the following guidelines:
\begin{itemize}
  \item \textbf{9--10 (Excellent overall quality)}:
    reasoning is highly correct (no major factual or spatial mistakes, no hallucinations,
    strongly aligned with the correct region) and highly useful (explicitly ties language
    to visual evidence, considers candidates, compares positions, and explains why the
    final region is chosen). This is the kind of CoT you would want as a
    \textbf{gold standard teaching signal}.
  \item \textbf{7--8 (Good overall quality)}:
    reasoning is mostly correct, with at most minor inaccuracies or slight vagueness.
    It is useful but may miss some steps or be less detailed (e.g., fewer explicit
    candidate comparisons). Still clearly grounded and helpful.
  \item \textbf{4--6 (Mixed / moderate quality)}:
    reasoning is partially correct (right general area or object type) but has notable
    gaps, imprecision, or some incorrect statements. Usefulness is limited: some
    connection between question and image exists, but the explanation is shallow,
    generic, or incomplete. This range also includes cases where the CoT is well
    structured but correctness is only moderate.
  \item \textbf{1--3 (Poor overall quality)}:
    reasoning is mostly incorrect or hallucinated, focusing on the wrong object or
    contradicting obvious visual evidence, or is so vague/generic that it provides
    almost no real grounding. Even if the final predicted bbox happens to be correct,
    if the CoT is wrong or useless, the score must remain low.
\end{itemize}

\medskip\hrule\medskip

\textbf{3.\ Evaluation Procedure}

For each sample:
\begin{enumerate}
  \item \textbf{Understand the target:}
    read the question to understand what region is being referred to (e.g., ``left-most ship'',
    ``top-center soccer field'', ``vehicle at bottom-right'', ``bridge in the middle'').
  \item \textbf{Inspect ground-truth and prediction:}
    use \texttt{ground\_truth} and \texttt{extracted\_answer} to understand which
    region is correct and which region was chosen by the model. This helps you judge
    if the reasoning aligns with the proper region.
  \item \textbf{Read the CoT inside \texttt{<think>...</think>}:}
    ignore anything outside \texttt{<think>...</think>} for scoring.
    Evaluate how correct and how useful this reasoning is as described above.
  \item \textbf{Assign a single overall score (1--10):}
    combine correctness and usefulness as described in Section~2.
    Also provide a brief explanation of why you chose this score.
\end{enumerate}

\medskip\hrule\medskip

\textbf{4.\ Output Format}

Return your evaluation as a \textbf{Python-style dictionary literal}, with no extra text:

\begin{flushleft}
\texttt{\{}\\
\quad \texttt{"score": <integer 1-10>,}\\
\quad \texttt{"explanation": "<1-3 short sentences explaining your judgment of correctness and usefulness together>"}\\
\texttt{\}}
\end{flushleft}

\end{tcolorbox}

\section{Training Configuration}
\label{sec:training_config}

We list the main GRPO training configurations for REC/OVD/GRES tasks in Table \ref{tab:rec_ovd_gres_config}. Different configs are listed first, while shared configs are merged across columns.

\begin{table*}[t]
\centering
\caption{Main GRPO training configurations for REC/OVD/GRES tasks.}
\label{tab:rec_ovd_gres_config}
\small
\setlength{\tabcolsep}{6pt}
\renewcommand{\arraystretch}{1.15}
\begin{tabularx}{\textwidth}{lXXX}
\toprule
\textbf{Config} & \textbf{REC (VRSBench-FS)} & \textbf{OVD (NWPU-FS)} & \textbf{GRES (EarthReason-FS)} \\
\midrule

\multicolumn{4}{l}{\textbf{Task-specific (different) configs}} \\
\midrule
Reward Name
& \textbf{\texttt{IoU}}
& \textbf{\texttt{mAP}}
& \textbf{\texttt{MaskGIoU}} \\

Per-device batch size
& 8
& 8
& 4 \\

Grad.\ accumulation
& 2
& 2
& 4 \\

\midrule
\multicolumn{4}{l}{\textbf{Shared (same) configs}} \\
\midrule
\#GPU
& \multicolumn{3}{c}{8} \\


Training budget
& \multicolumn{3}{c}{num\_train\_epochs=30} \\

Rollout Size
& \multicolumn{3}{c}{num\_generations=8} \\

Max completion length
& \multicolumn{3}{c}{2048} \\

Learning rate
& \multicolumn{3}{c}{\emph{1.0e-6}} \\

Weight decay
& \multicolumn{3}{c}{\emph{1.0e-2}} \\

Seed
& \multicolumn{3}{c}{data\_seed=42} \\

Reward funcs
& \multicolumn{3}{c}{\texttt{metrics} + \texttt{format}} \\

GRPO $\beta$
& \multicolumn{3}{c}{0.04} \\



Optimizer
& \multicolumn{3}{c}{AdamW} \\

\bottomrule
\end{tabularx}
\end{table*}

\section{More experimental results}

\subsection{Variance and Robustness}
The few-shot training samples were randomly selected. To further assess robustness with respect to shot selection, we re-sampled the training data using different random seeds and re-ran the experiments for the REC and GRES tasks. For OVD, we report 5-shot result (10-shot setting for OVD is hard to re-sample different samples). As shown in the Table~\ref{tab:shuffle}, our model exhibits stable performance across 10 different seeds, with standard deviations below 1\%. 

\begin{table*}
    \centering
    \caption{Robustness with Different Few-shot Samples.}
    \resizebox{\textwidth}{!}{%
    \begin{tabular}{l|c|c|c|c|c|c|c|c|c|c|c|c|c}
        \midrule
        \textbf{Task} & \textbf{Metrics} & \textbf{Reported} (Seed 42) & \textbf{Average $\pm$ std} & Seed 43 & Seed 44 & Seed 45 & Seed 46 & Seed 47 & Seed 48 & Seed 49 & Seed 50 & Seed 51 & Seed 52 \\
        \midrule
        REC       & Acc@0.5 & 50.59 & $49.46 \pm 0.95$ & 49.90 & 49.76 & 50.34 & 49.77 & 50.28 & 47.43 & 49.20 & 49.61 & 50.33 & 47.99 \\
        \midrule
        OVD	& mAP	&33.20	&$33.68 \pm 0.97$	&34.36	&33.71	&34.79	&33.50	&32.10	&34.20	&34.74	&32.09	&32.87	&34.48	\\
        \midrule
        GRES-Val  & gIoU    & 57.78 & $59.19 \pm 0.31$ & 59.08 & 58.78 & 59.41 & 59.58 & 59.57 & 58.80 & 59.33 & 59.52 & 58.89 & 58.94 \\
        \midrule
        GRES-Test & gIoU    & 58.27 & $58.05 \pm 0.63$ & 59.00 & 58.20 & 58.18 & 57.27 & 58.41 & 57.10 & 58.47 & 58.59 & 58.17 & 57.11 \\
        \bottomrule
    \end{tabular}%
    }
    \label{tab:shuffle}
\end{table*}

\subsection{Assessment on Generated Reasoning Chains}
\label{sec:assess}
\begin{figure}
\begin{center}
\includegraphics[width=0.7\columnwidth]{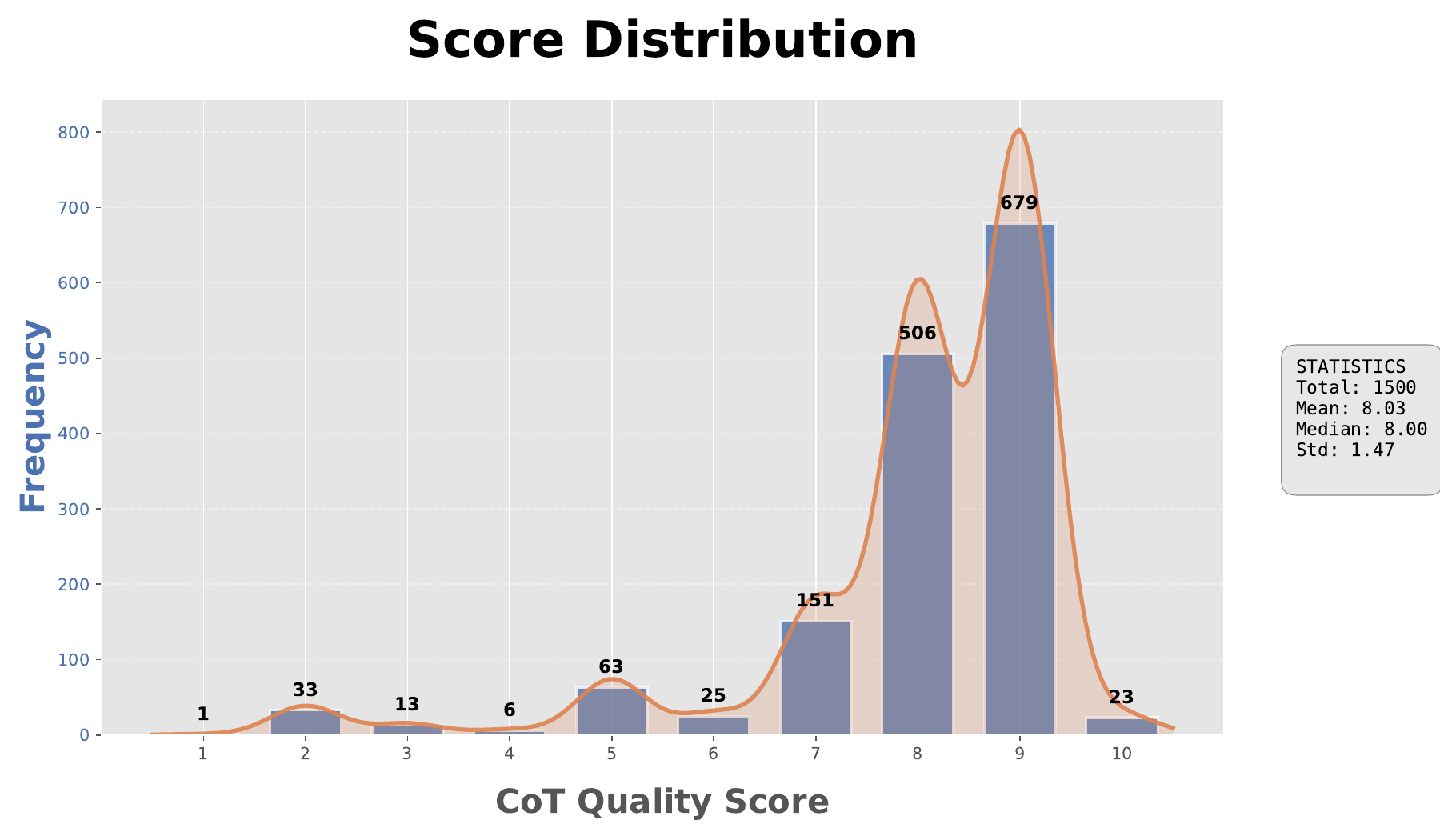}
\end{center}
\caption{Statistics of Quality Assessment on Generated Reasoning Chains.}
\label{fig:cotassess}
\end{figure}

To further validate reasoning quality, we conducted an additional analysis in which we prompted Qwen3-VL-235B-A22B-Thinking~\citep{bai2025qwen3vltechnicalreport} to automatically evaluate a randomly sampled subset of reasoning chains whose corresponding predictions have IoU $>$ 0.5 on the VRSBench test set (1500 samples). We carefully designed prompts to assess both the correctness and usefulness of the reasoning toward the final prediction, assigning score on a 1–10 scale. As summarized in the Fig.~\ref{fig:cotassess}, the average reasoning quality score is 8.03 with std of 1.47, indicating that the majority of generated reasoning chains are reasonable, informative, and supportive of the final predictions.

\subsection{Failure Case Analysis}
\label{sec:failure}
\begin{figure*}
\begin{center}
\includegraphics[width=0.95\textwidth]{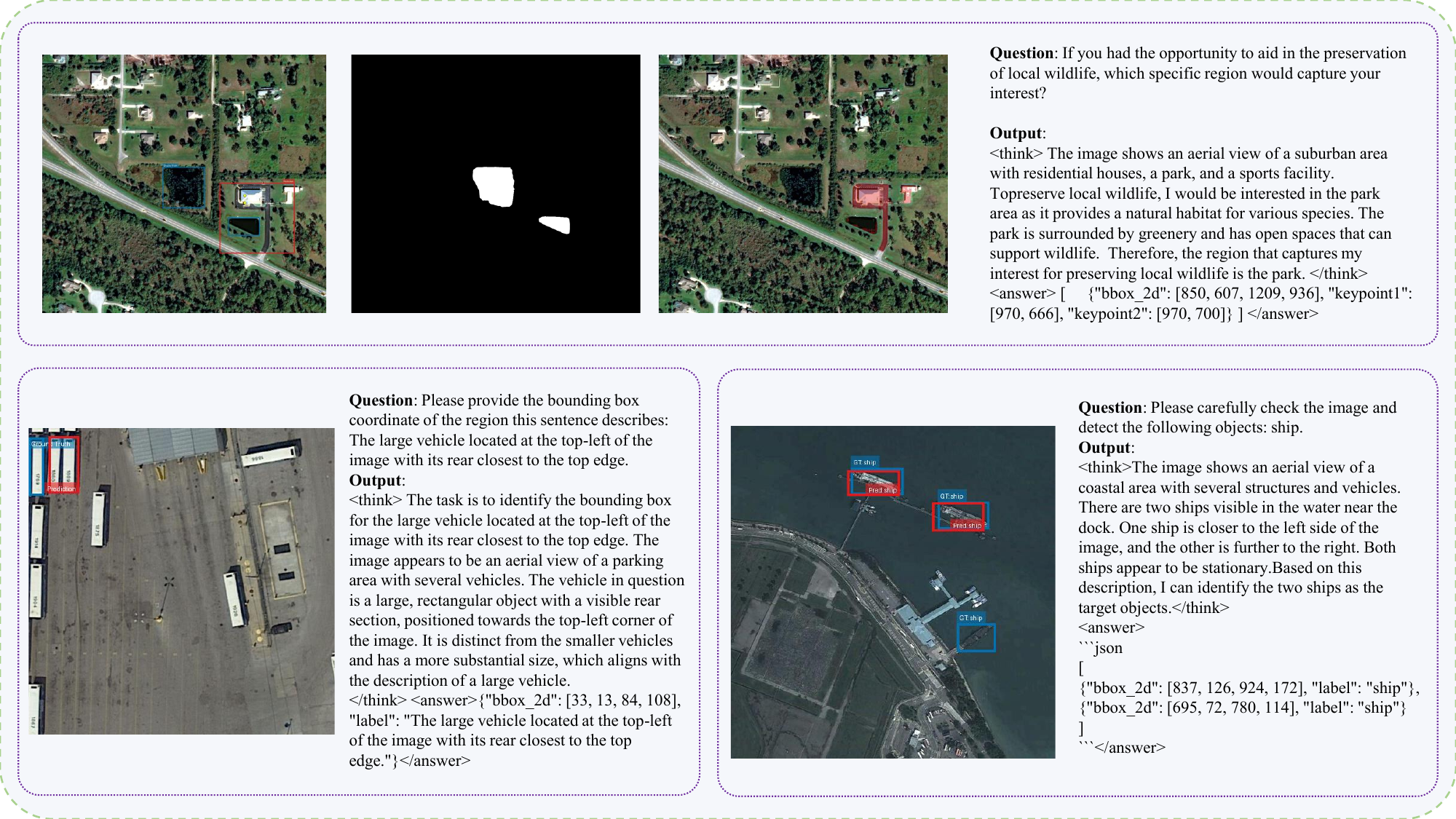}
\end{center}
\caption{Geo-R1 inference samples (failure case).}
\label{fig:samples3}
\end{figure*}

\textbf{Qualitative Analysis.} \textcolor{black}{As shown in the top of Fig.~\ref{fig:samples3}, the model can provide meaningful reasoning chain. Park is the most suitable region for wildlife preservation, yet its final answer incorrectly grounds this concept to the nearby sports facility. In the bottom-right example, the model's reasoning incorrectly identifies the presence of multiple ships, resulting the final answer detecting only the cluster on the left while omitting the group on the right. Furthermore, we also observe failures in reasoning process, the bottom-left example shows a failure within the reasoning process itself. Although the model accurately understands the task in its reasoning block (to identify the `large vehicle'), it subsequently fails to apply this critical size attribute, incorrectly selecting a smaller, adjacent vehicle for its final answer. These cases suggest that a key area for future improvement is strengthening the alignment between the model's high-level semantic understanding and the generation of accurate, complete, and precise spatial coordinates.}

\textbf{Quantitative Analysis.} \textcolor{black}{To provide a comprehensive view of the failure distribution, we conducted a manual statistical analysis on the failure cases across OVD ($N=50$), REC ($N=500$), and RES ($N=500$) tasks. We categorize the errors into three primary types: 1) \textit{Localization Shift}, representing precise localization failures where the predicted box/mask drifts from the ground truth; 2) \textit{Object Confusion}, where the model incorrectly targets a semantic distractor instead of the target; and 3) \textit{Spatial Relation Error}, where the model misinterprets relative positional descriptions. As reported in Table~\ref{tab:failure_stats}, \textbf{Localization Shift} constitutes the dominant error mode across all tasks, particularly peaking in OVD (52.0\%) and RES (47.4\%). This suggests that while the model often locates the correct general region, precise boundary regression remains a bottleneck. \textbf{Object Confusion} is notably more prevalent in REC (35.5\%) and RES (31.6\%) compared to OVD (28.0\%), indicating that resolving linguistic ambiguity in referring expressions is more prone to semantic interference than category-based detection. \textbf{Spatial Relation} errors consistently account for approximately 20\%--24\% of failures across all tasks, highlighting a persistent challenge in geometric reasoning.}


\begin{table}[!h]
    \centering
    \color{black}
    \caption{\textcolor{black}{Statistics of failure case types across OVD, REC, and RES tasks.}}
    \label{tab:failure_stats}
    \resizebox{0.6\columnwidth}{!}{%
    \begin{tabular}{l|cc|cc|cc}
        \toprule
         \multirow{2}{*}{\textbf{Error Type}} & \multicolumn{2}{c|}{\textbf{OVD} ($N=50$)} & \multicolumn{2}{c|}{\textbf{REC} ($N=500$)} & \multicolumn{2}{c}{\textbf{RES} ($N=500$)} \\
         & Count & Ratio & Count & Ratio & Count & Ratio \\
        \midrule
        Localization Shift & 26 & \textbf{52.0\%} & 204 & \textbf{40.8\%} & 237 & \textbf{47.4\%} \\
        Object Confusion & 14 & 28.0\% & 178 & 35.5\% & 158 & 31.6\% \\
        Spatial Relation Error & 10 & 20.0\% & 118 & 23.7\% & 105 & 21.1\% \\
        \midrule
        \textbf{Total} & 50 & 100\% & 500 & 100\% & 500 & 100\% \\
        \bottomrule
    \end{tabular}%
    }
\end{table}

\section{DPO Training Dataset Construction}
\label{sec:dpo_construct}

\textcolor{black}{A faithful DPO baseline is non-trivial in our REU setting since it requires constructing paired preference data (chosen/rejected) at scale, which is not readily available for remote-sensing tasks. Take REC as an example, to train a DPO baseline for REC task, we construct a DPO dataset $\mathcal{D}_{\text{dpo}}$.}

\textcolor{black}{For each image-question pair $(I, Q)$, we first use the Qwen2.5-VL-72B teacher policy $\pi_{\text{teacher}}$ to generate $N$ rollouts $\{s_i\}_{i=1}^{N}$, where each rollout $s_i$ contains a chain-of-thought reasoning path $r_i$ and a predicted bounding box $b_i$. Specifically,
\begin{equation}
\mathcal{O} = \{s_i\}_{i=1}^{N} = \{(r_i, b_i)\}_{i=1}^{N} \sim \pi_{\text{teacher}}(\cdot \mid I, Q).
\label{eq:teacher_rollouts}
\end{equation}
}

\textcolor{black}{We then compute the IoU score $k_i = \text{IoU}(b_i, b_{\text{gt}})$ between each predicted box $b_i$ and the ground-truth box $b_{\text{gt}}$.}

\textcolor{black}{Instead of random negative sampling, we select hard negatives by choosing rollouts whose IoU falls into a \emph{confusion interval} $[\tau_{\min}, \tau_{\max}]$. These rollouts become the rejected response $y_{\text{rej}}$.
This design encourages learning from boundary cases rather than trivial failures.
For the chosen response $y_{\text{chs}}$, we use a high-quality rollout (high IoU) together with its reasoning path. Overall, the constructed preference dataset consists of multiple chosen--rejected pairs:
\begin{equation}
\mathcal{D}_{\text{DPO}} = \{(y_{\text{chs}}, y_{\text{rej}})\}.
\label{eq:dpo_dataset}
\end{equation}
}

\textcolor{black}{
In particular, the chosen response is defined as
\begin{equation}
y_{\text{chs}} = (r_m, b_m), \exists\, m \in \{1,\ldots,N\}\ \text{s.t.}\ k_m > \tau_{\text{high}},
\label{eq:y_chosen}
\end{equation}
and the rejected response is defined as
\begin{equation}
y_{\text{rej}} = (r_n, b_n), \exists\, n \in \{1,\ldots,N\}\ \text{s.t.}\ \tau_{\min} < k_n < \tau_{\max}.
\label{eq:y_reject}
\end{equation}
}

\textcolor{black}{
To ensure sufficient distinguishability between the chosen and rejected responses, we further enforce an IoU margin constraint that must be satisfied simultaneously:
\begin{equation}
\delta_{\min} < k_m - k_n < \delta_{\max}.
\label{eq:iou_margin}
\end{equation}
}

\textcolor{black}{
In our implementation, we set $\tau_{\text{high}} = 0.65$ for selecting the chosen response. For hard-negative mining, we use $\tau_{\min} = 0.3$ and $\tau_{\max} = 0.5$. The IoU margin is set to range from $\delta_{\min} = 0.2$ to $\delta_{\max} = 0.45$.
}

\section{Reward and Context Length Stability during Training}
\label{sec:reward_contextlen}

\textcolor{black}{Across all three tasks, we observe consistent training dynamics in terms of reward learning and generation length control.
First, the \textbf{format reward is learned very quickly}. In REC/OVD/GRES, the model rapidly acquires the required output schema and then maintains a consistently high format reward throughout training, indicating strong instruction-following and stable formatting behavior.
Second, the \textbf{convergence speed differs across metric rewards}: the mAP reward in OVD typically converges within the first few tens of steps with relatively small oscillations, while the IoU reward in REC converges more gradually. The GIoU reward in GRES converges the slowest and exhibits the strongest oscillations, which is likely related to the higher variance and increased difficulty of mask-level reward signal.
Third, the \textbf{output length remains well controlled}. The MA10 token length quickly stabilizes into a bounded range for each task, suggesting that GRPO does not induce uncontrolled response lengthening and that the generation behavior remains stable during optimization.}

\subsection{REC}
 
\begin{figure}[h]
    \centering
    \begin{subfigure}[t]{0.49\linewidth}
        \centering
        \includegraphics[width=\linewidth]{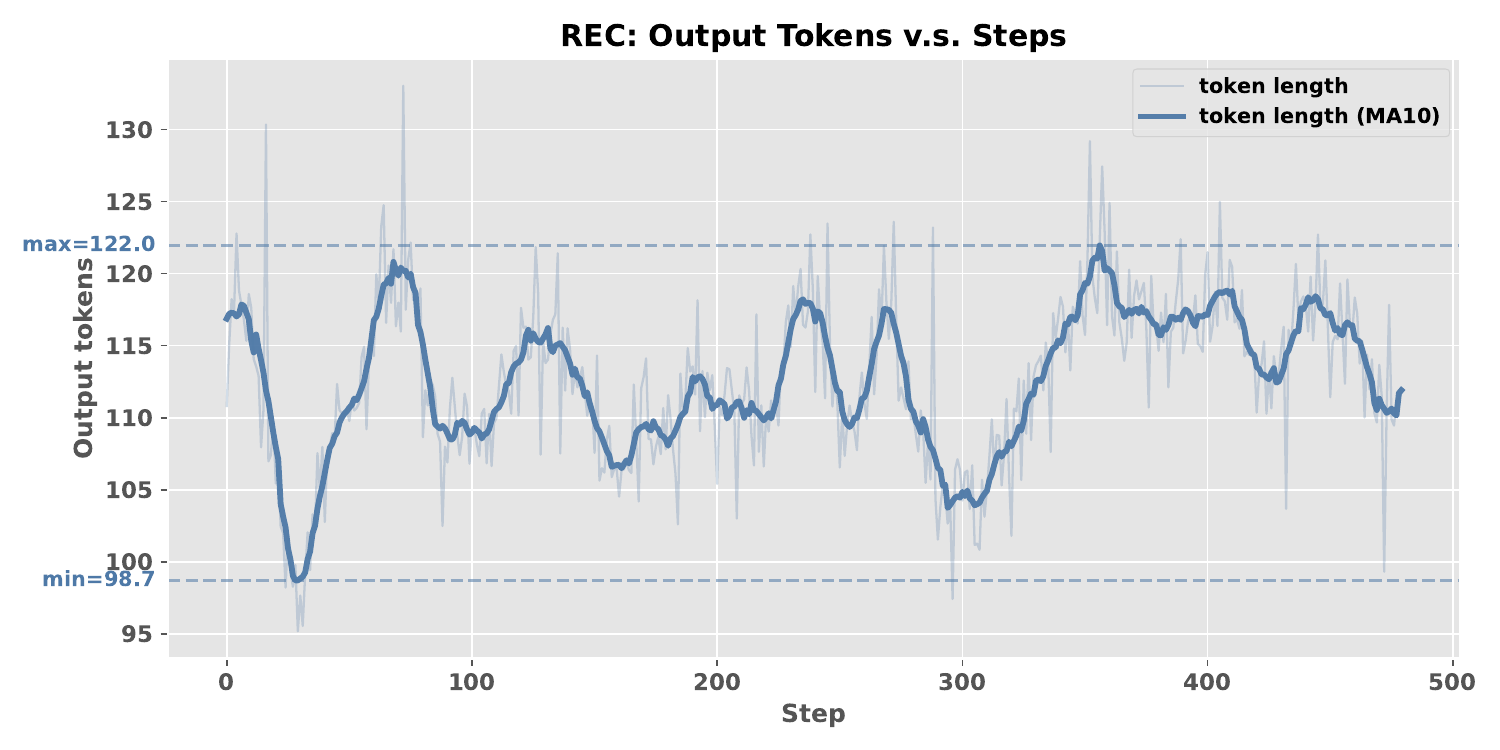}
        \caption{REC: output token length vs.\ training steps. The MA10 remains stable (min=98, max=122).}
        \label{fig:rec_tokens_vs_steps}
    \end{subfigure}
    \hfill
    \begin{subfigure}[t]{0.49\linewidth}
        \centering
        \includegraphics[width=\linewidth]{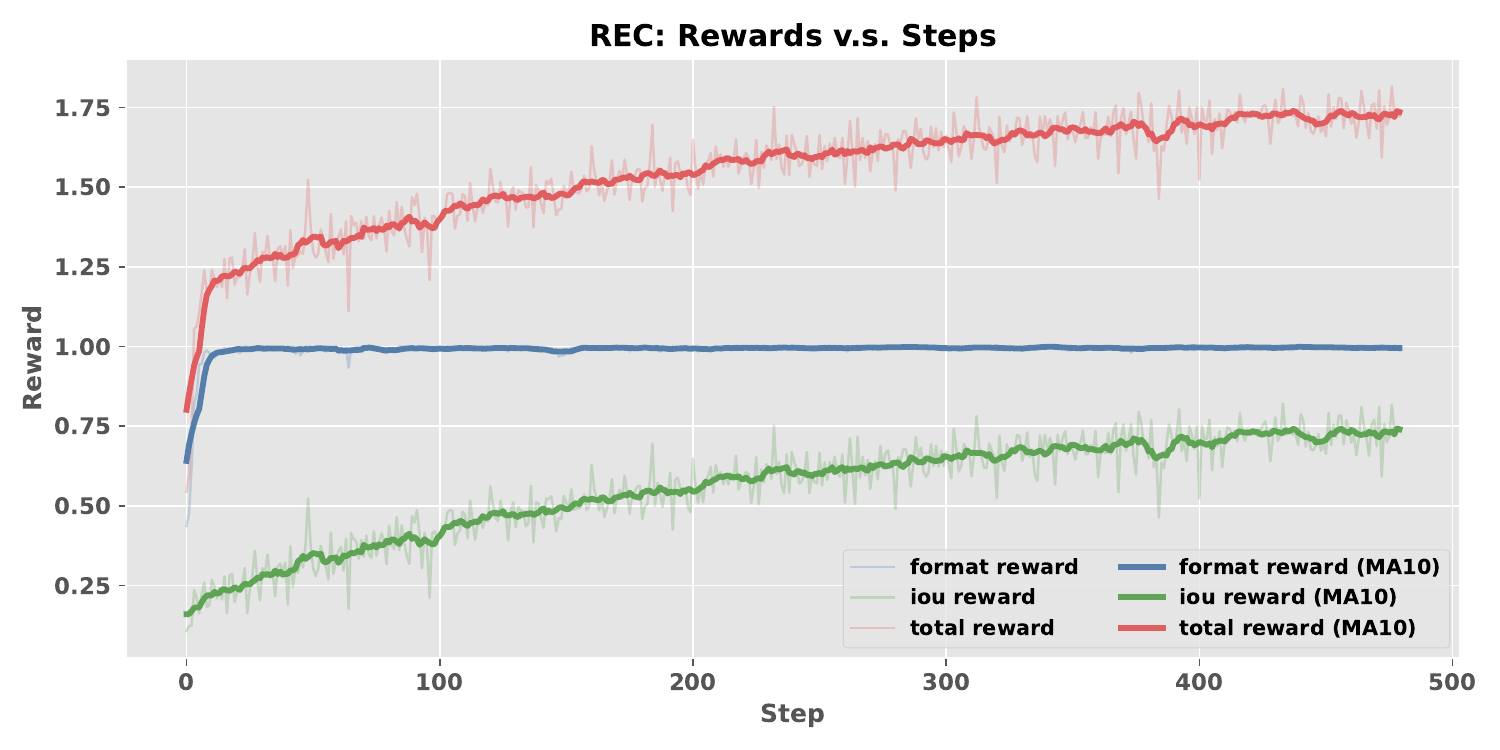}
        \caption{REC: reward curves vs.\ training steps (format/IoU/total), with MA10 smoothing. Rewards improve smoothly and converge without collapse.}
        \label{fig:rec_reward_vs_steps}
    \end{subfigure}
    \caption{REC training stability analysis: (a) output token length; (b) reward evolution.}
    \label{fig:rec_stability_two_plots}
\end{figure}

\textcolor{black}{To empirically verify length stability, we track the \textbf{output token length} during REC GRPO training and plot the MA10 (moving average, window size 10).
As shown in Fig.~\ref{fig:rec_tokens_vs_steps}, the MA10 of response length stabilizes within a narrow range (\textbf{min=98, max=122} tokens), indicating that our training does not induce uncontrolled generation lengthening.}

\textcolor{black}{We further report the evolution of rewards during training.
Fig.~\ref{fig:rec_reward_vs_steps} plots the format reward, IoU reward, and total reward (with MA10 smoothing). The curves show a smooth improvement and convergence without collapse, and both the task metric reward (IoU) and total reward increase steadily.
This indicates that the GRPO optimization is stable in our experiments.}

\subsection{OVD}

\begin{figure}[h]
    \centering
    \begin{subfigure}[t]{0.49\linewidth}
        \centering
        \includegraphics[width=\linewidth]{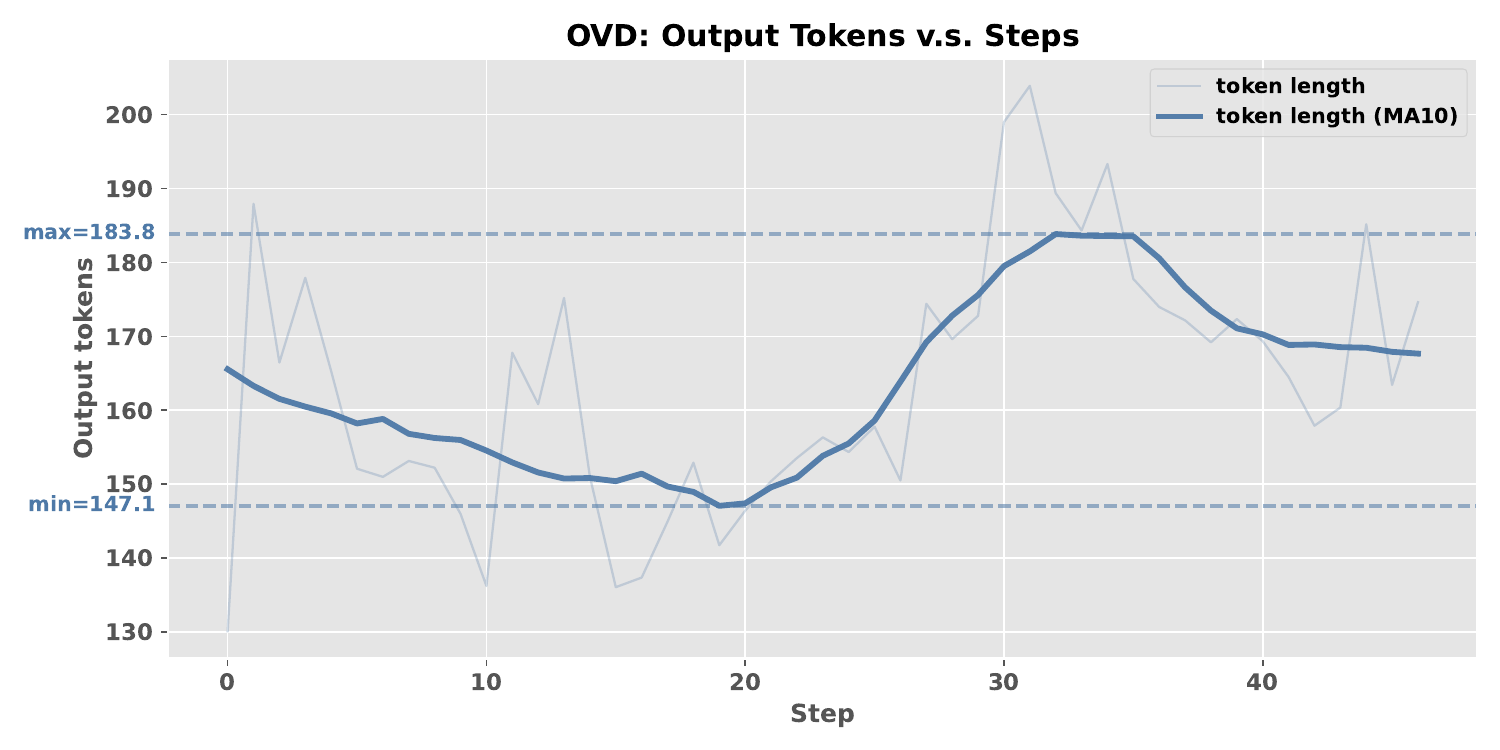}
        \caption{OVD: output token length vs.\ training steps (MA10). min=147, max=183.}
        \label{fig:ovd_tokens_vs_steps}
    \end{subfigure}
    \hfill
    \begin{subfigure}[t]{0.49\linewidth}
        \centering
        \includegraphics[width=\linewidth]{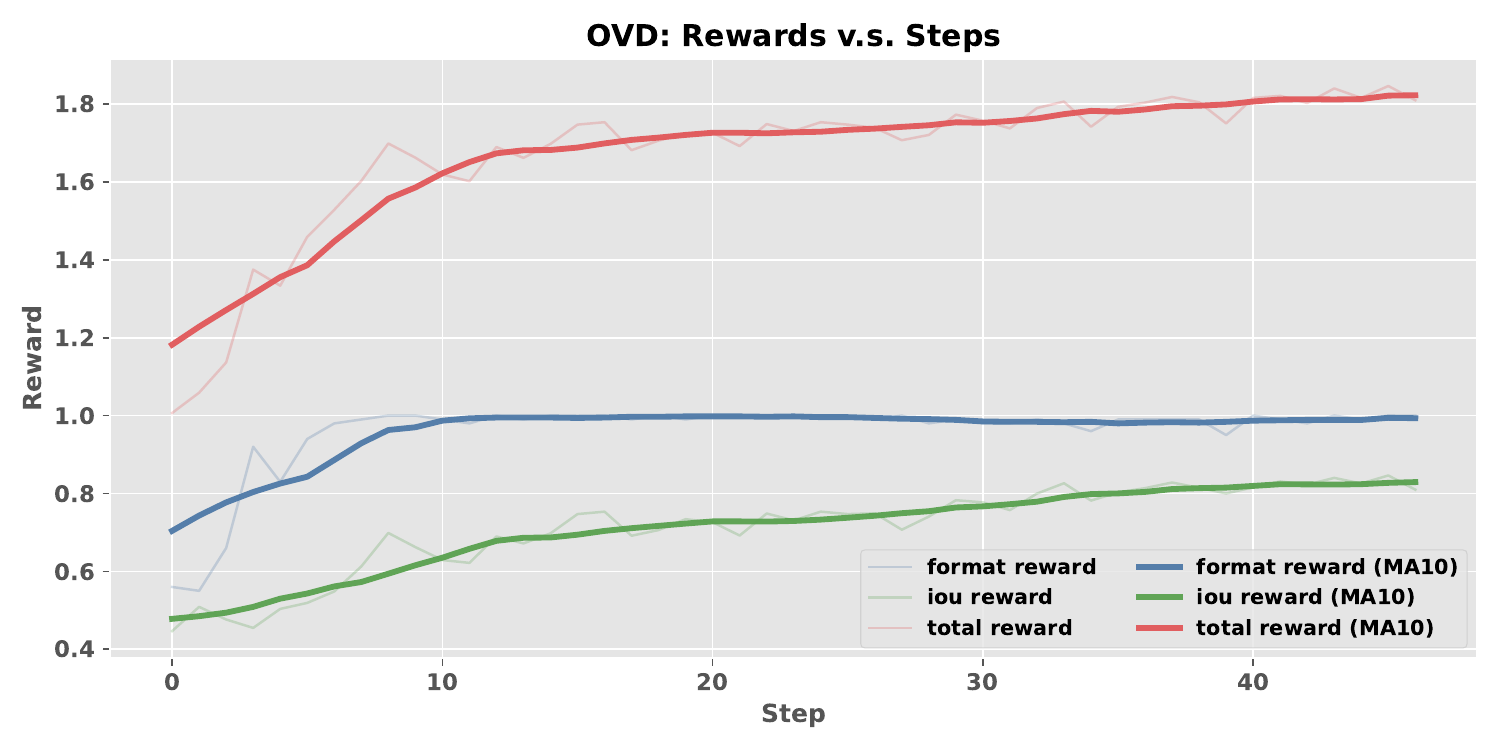}
        \caption{OVD: reward curves vs.\ training steps (format/mAP/total), with MA10 smoothing.}
        \label{fig:ovd_reward_vs_steps}
    \end{subfigure}
    \caption{OVD training stability analysis: (left) output token length; (right) reward evolution.}
    \label{fig:ovd_stability_two_plots}
\end{figure}

\textcolor{black}{We conduct the same stability analysis for OVD. We track the output token length during OVD GRPO training and plot the MA10 trend.
As shown in Fig.~\ref{fig:ovd_tokens_vs_steps}, the response length remains stable within a bounded range (\textbf{min=147, max=183} tokens). The \textbf{longer response length compared with the REC task} is expected, since OVD may require predicting multiple targets in a single image. }

\textcolor{black}{We also visualize reward evolution in Fig.~\ref{fig:ovd_reward_vs_steps}, including the format reward, mAP reward, and total reward (with MA10 smoothing).
The reward curves exhibit smooth progress without collapse, indicating stable optimization. We further observe that the mAP-related reward in OVD converges rapidly within the first few tens of steps, and exhibits \textbf{smaller oscillations} than the IoU reward in REC. We hypothesize this faster and smoother convergence is partly due to the \textbf{lower effective image diversity} in our OVD few-shot setting (a single image can yield multiple few-shot samples and may reduce feature variety compared with REC).}

\subsection{GRES}

\begin{figure}[h]
    \centering
    \begin{subfigure}[t]{0.49\linewidth}
        \centering
        \includegraphics[width=\linewidth]{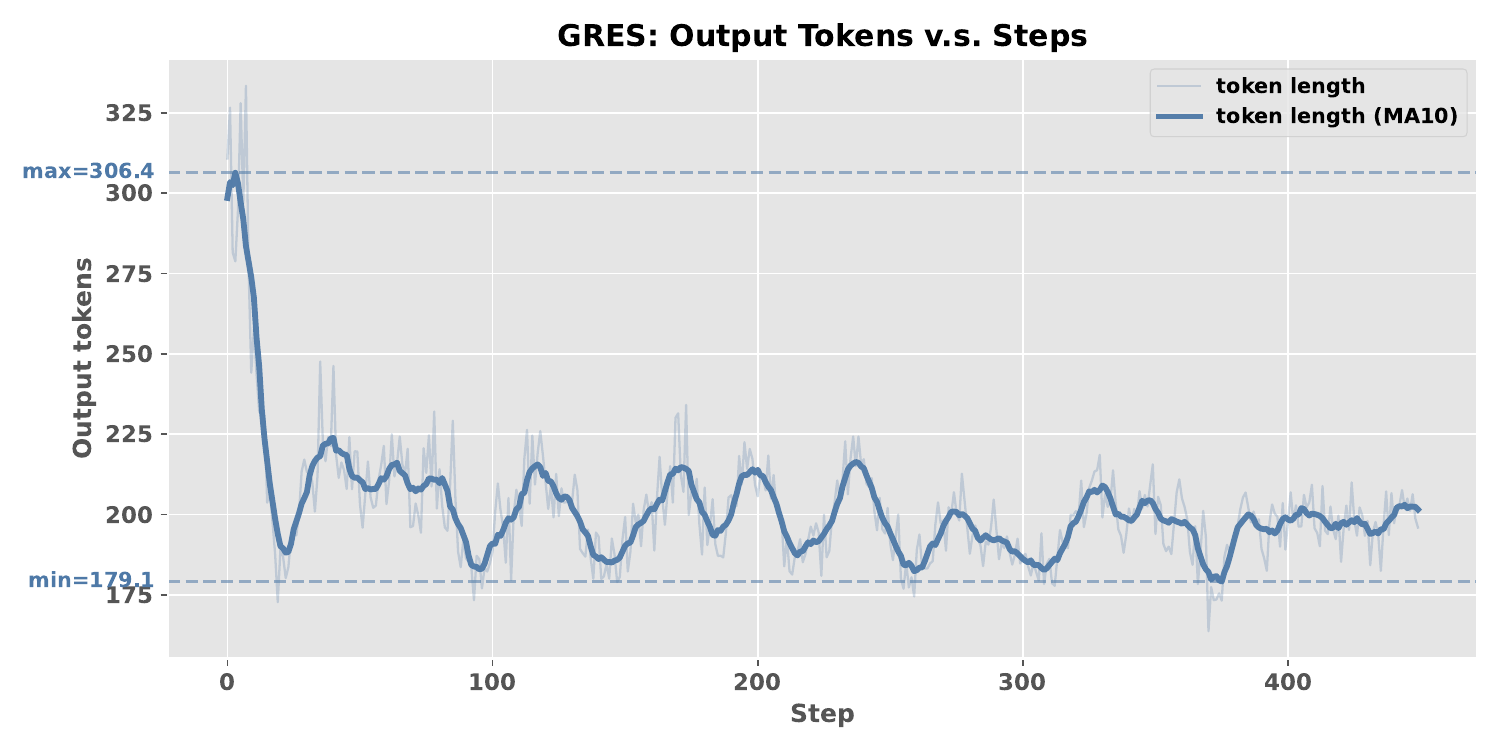}
        \caption{GRES: output token length vs.\ training steps (MA10). min=179, max=306.}
        \label{fig:gres_tokens_vs_steps}
    \end{subfigure}
    \hfill
    \begin{subfigure}[t]{0.49\linewidth}
        \centering
        \includegraphics[width=\linewidth]{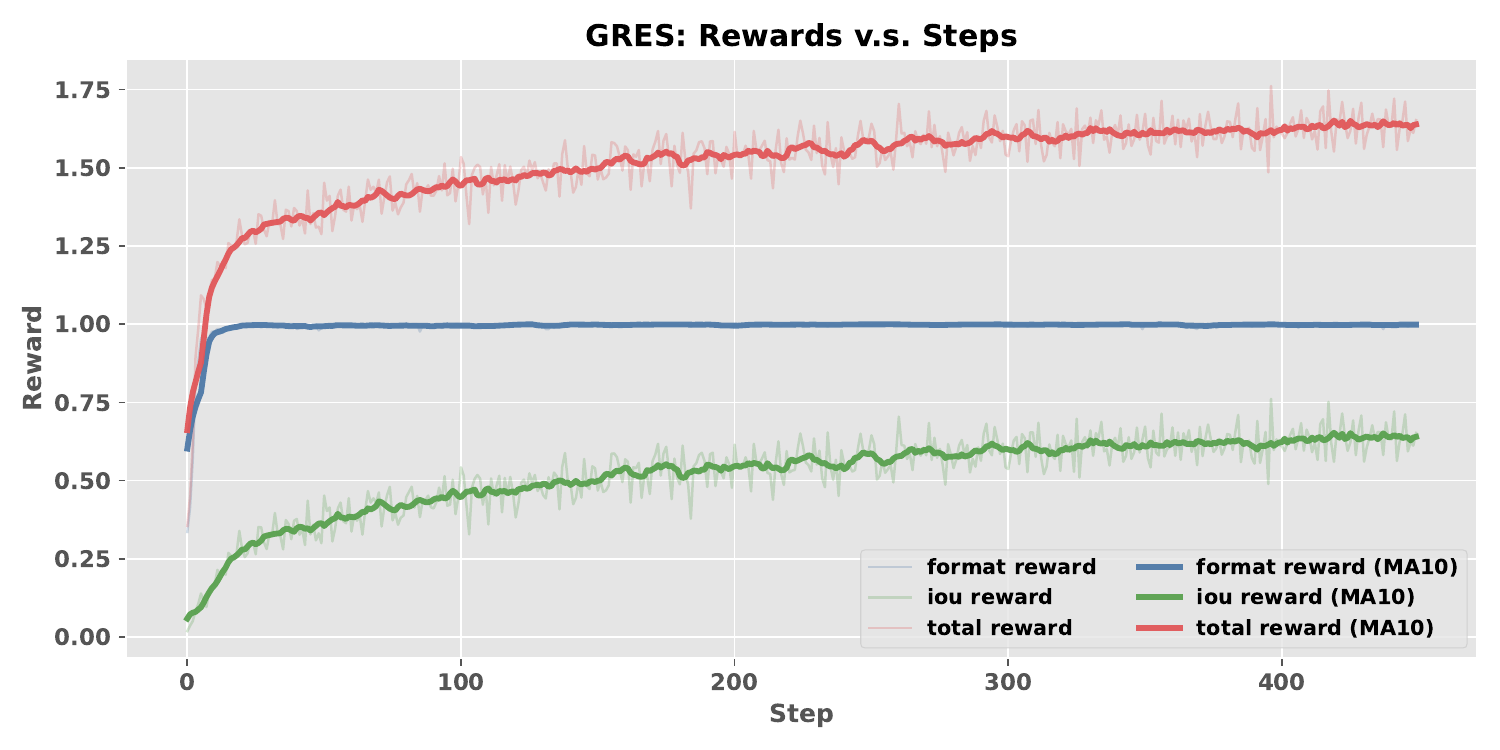}
        \caption{GRES: reward curves vs.\ training steps (format/gIoU/total), with MA10 smoothing.}
        \label{fig:gres_reward_vs_steps}
    \end{subfigure}
    \caption{GRES training stability analysis: (left) output token length; (right) reward evolution.}
    \label{fig:gres_stability_two_plots}
\end{figure}

\textcolor{black}{We perform the same stability analysis for GRES. We track the \textbf{output token length} during GRES GRPO training and plot the MA10 trend.
As shown in Fig.~\ref{fig:gres_tokens_vs_steps}, the MA10 token length \textbf{quickly converges within the first few tens of steps} and then remains in a stable range of \textbf{179-306} tokens. }

\textcolor{black}{We further visualize reward trend in Fig.~\ref{fig:gres_reward_vs_steps}, including the format reward, MaskGIoU reward, and total reward (with MA10 smoothing).
The curves show that the gIoU-related reward and total reward increase steadily while the format reward remains consistently high after few tens of steps, reflecting stable optimization and instruction-follow to the output schema throughout training. Notably, the convergence of GRES is \textbf{slower} than that of REC and OVD (GREC), likely due to the increased difficulty of higher variance introduced by mask-level reward.}

\end{document}